\pdfoutput=1
\documentclass[11pt]{article}

\usepackage{acl}

\usepackage[T1]{fontenc}
\usepackage[utf8]{inputenc}

\usepackage{microtype}

\usepackage{inconsolata}

\usepackage{times}
\usepackage{tabu}
\usepackage{latexsym}
\usepackage{stmaryrd}
\usepackage{amsmath}
\usepackage{multirow}
\usepackage{bbm}
\usepackage{graphicx}
\usepackage{amssymb}
\usepackage{booktabs}
\usepackage{adjustbox}
\usepackage{xspace}
\usepackage{algpseudocode}
\usepackage{algorithm}
\usepackage{graphicx}
\usepackage{caption}
\usepackage{subcaption}
\usepackage{tabularx}
\usepackage{colortbl}
\usepackage{xcolor}
\usepackage{tablefootnote}
\usepackage{enumitem}  % More control
\usepackage{lipsum}
\usepackage{xspace}
\usepackage{bbding}
\usepackage{pifont}
\usepackage{arydshln}
\usepackage{array}
\usepackage{todonotes}
\usepackage{comment}
\usepackage{listings} % code snippets

\usepackage{cleveref}
\usepackage{placeins}

\definecolor{sociodemographics}{rgb}{0.0, 0.55, 0.55}
\definecolor{prompt}{rgb}{0.8, 0.36, 0.27}
\definecolor{dataset}{rgb}{0.72, 0.53, 0.04}
\newcommand{\colorline}[2]{{\color{#1}#2}}
\definecolor{tablegray}{rgb}{0.2,0.2,0.2}

\title{Contextualized Prompting For \\ Stance Detection On Social Media}

\author{Tilman Beck$^{1}$ \qquad Shakib Yazdani$^{2}$ \qquad Simon Kruschinski$^{3}$ \\ \qquad \textbf{Marcus Maurer}$^{4}$ \qquad \textbf{Iryna Gurevych}$^{5}$\\
	$^1$Institute of Intensive Care, University Hospital of Zurich and University of Zurich \\
        $^{2}$Institute of Computer Science, University of Goettingen \\
        $^{3}$GESIS Leibniz Institute for the Social Sciences \\
        $^{4}$Institut f{\"u}r Publizistik, Johannes Gutenberg-University Mainz \\
        $^{5}$Ubiquitous Knowledge Processing Lab, Technical University of Darmstadt \\
	\texttt{beck.tilman@gmail.com} \\
}
\begin{document}
\maketitle
\begin{abstract}
Stance detection on social media is challenging due to short, noisy, and context-dependent language. 
While large language models (LLMs) show zero-shot generalization, they are typically prompted without contextual information, which limits their ability to interpret ambiguous posts. 
In this work, we systematically investigate the impact of incorporating real-world (e.g., user biographies), derived (e.g., political party), and LLM-generated (e.g., target descriptions) contextual features into zero-shot prompting for stance detection on Twitter. 
Our evaluation spans four benchmark datasets, including a new high-quality German Twitter stance dataset.
Across multiple LLMs, we find that integrating contextual information improves performance, but only under specific conditions. 
LLM-generated target descriptions consistently enhance accuracy, while other user metadata has mixed or even detrimental effects. 
Notably, we show that the inclusion of other tweets by the same user, often beneficial in supervised learning, can impair performance due to input noise. 
Our qualitative analysis reveals that LLMs struggle to distinguish task-specific useful information from irrelevant context. 
Our findings highlight both the promise and challenges of prompting with context information in noisy real-world settings. 
We publish code and data at this \href{https://github.com/tilmanbeck/stance-context-twitter}{page}.
\end{abstract}

\section{Introduction}

Given a text, stance detection is the task of identifying an author's viewpoint towards a specific target, such as an individual~\citep{sobhani-etal-2017-dataset}, a controversial topic~\citep{murakami-raymond-2010-support, mohammad-etal-2016-dataset}, or a product~\citep{somasundaran-wiebe-2009-recognizing}.
Although LLMs have shown promising zero-shot and few-shot generalization capabilities in stance detection~\citep{taranukhin-etal-2024-stance}, social media remains a challenging domain.
Posts are often short, informal, noisy and highly context dependent~\citep{socialmedia_challenges2018}, and a single tweet may lack sufficient information to provide a complete understanding of the author's stance on a topic.
However, the standard setup for stance detection typically involves classifying tweets in isolation, without considering additional context~\citep{schiller2021benchmark, hardalov-etal-2021-cross}.

\begin{figure}[h]
	\centering
	\includegraphics[scale=0.7]{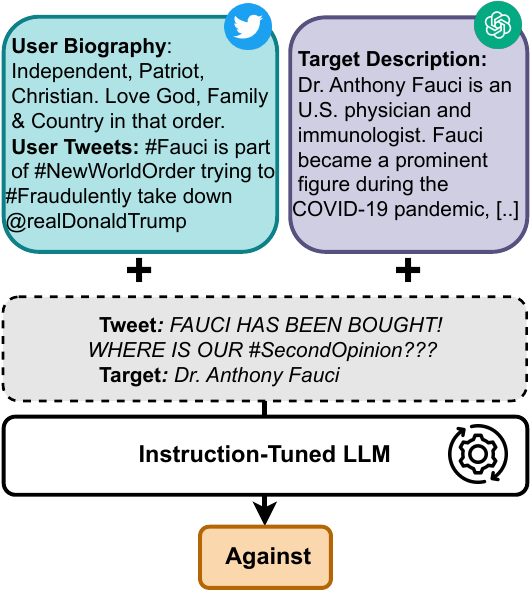}
	\caption{Additional context information such as author metadata from Twitter (cyan) or an LLM-generated target description (lavender) can provide additional cues for interpreting the stance of Twitter posts towards a specific target (grey). We investigate the influence on model performance in stance detection when adding such contextual features to the prompt.}
	\label{fig:context_prompting}
\end{figure}

Various types of additional information can offer valuable clues for interpreting ambiguous expressions (see \Cref{fig:context_prompting} for an example).
We refer to this information as \textit{contextual features}.
Specifically, Twitter metadata, such as user biographies and other tweets from the same user, often contain relevant information to predict the stance of a user~\citep{burfoot-etal-2011-collective, sridhar-etal-2015-joint, porco-goldwasser-2020-predicting, stem2022}. 
Several works suggest that incorporating such metadata into supervised model training improves stance detection performance~\citep{hasan-ng-2013-stance, li-etal-2018-structured, samih-darwish-2021-topical}, especially when tweets are ambiguous.
However, it is unclear whether these positive effects translate to prompting scenarios, particularly due to LLMs' susceptibility to prompt design~\citep{webson-pavlick-2022-prompt, min-etal-2022-rethinking, lu-etal-2022-fantastically}.

At the same time, LLMs are increasingly being used to generate or augment training data for NLP tasks~\citep{li-etal-2023-synthetic, liu2024bestpractices, li-etal-2024-empowering, zhao-etal-2024-zerostance}, e.g., by creating semantically distinct examples via few-shot prompting~\citep{chung-etal-2023-increasing, yu2023diversitybias}.
In contrast, extracting meaningful contextual features for existing data is less explored~\citep{taranukhin-etal-2024-stance}.
Yet, such context could, in theory, enable models to make more accurate predictions by supplying them with background knowledge about the topic being discussed or the user who authored the text.

Independent of their origin, contextual features potentially carry useful cues, but can also be irrelevant or even conflicting in some cases (e.g., a user who indicates a pro-vaccination stance in their biography but tweets against a specific vaccine).
Prior work on ignoring irrelevant context targets tasks where relevance is objectively defined, such as math word problems~\citep{wu-etal-2024-instructing} or knowledge-intensive tasks~\citep{wu2024}. 
No comprehensive evaluation exists for prompt-based classification of subjective tasks like stance detection.

%Despite these advancements, it remains unclear whether LLMs can meaningfully leverage contextual features in zero-shot prompting scenarios, due to their susceptibility to prompt design~\citep{webson-pavlick-2022-prompt, min-etal-2022-rethinking, lu-etal-2022-fantastically}.

We address this gap by investigating the impact of different contextual features on the performance of LLMs for prompt-based stance detection.
Specifically, we integrate both real (e.g., user biography), derived (e.g., political party derived from user biography), and LLM-generated contextual features into the prompt to assess their impact on model performance across four different Twitter stance detection benchmarks.

%%%%%  GAP %%%%%%
% what is the gap? why is it important? why is it challenging?
%Although the positive contribution of few-shot examples for prompting performance is well-established for many classification tasks in NLP~\citep{brown2020language, wei2022finetuned, sanh2022zero}, the impact of combining different context on model performance is less explored.
%In contrast to beforementioned work, we aim to integrate various sources of contextual information, such as metadata of user posts (e.g., user biography) and synthetically generated instance-specific text (e.g., prompting topic-relevant information).
%Such information potentially carries import cues for the interpretation of a tweet, but can also be irrelevant or even conflicting in some cases (e.g., a user who indicates a pro-vaccination stance in their biography but tweets against a specific vaccine).
%It is unclear whether LLMs can effectively leverage such context information and differentiate without explicit fine-tuning, as it is done in supervised settings.
%Further, we assess combinations of context features and their impact on the model performance in a zero-shot prompting setup for stance detection.
%This allows to investigate if there is a trade-off of the amount of context information and model performance.

%%%%%%%%%%%%%%%%%%%%%%%%
%%%% Contributions %%%%%
%%%%%%%%%%%%%%%%%%%%%%%%

% general insights: which context features are most/least informative for stance detection in Twitter
We summarize our contributions in the following; \textbf{(1)} we show that the integration of additional context features can improve model performance, but that the impact is dependent on the specific contextual feature and dataset.
While integrating LLM-generated target descriptions is always beneficial, incorporating other tweets from the same user into the prompt can have a negative impact on performance, which is in contrast to previous work on \textit{supervised} stance detection~\citep{samih-darwish-2021-topical};
% does this generalise across different datasets and models?
\textbf{(2)} we investigate whether the findings generalize across different families of reasoning-based LLMs (i.e., GPT-OSS, Gemma3, Qwen2.5, Ministral3).
Again, LLM-generated target descriptions are generally beneficial while the impact of other contextual features is less consistent across different model families and sizes.
\textbf{(3)} our analysis demonstrates that all contextual features potentially provide useful information but that discriminating between informative and irrelevant information is challenging for prompting using LLMs;
\textbf{(4)} finally, we publish a new high-quality (Krippendorff $\alpha$>0.67) Twitter stance detection dataset in German about governmental measures during Covid-19 (n=3,750).

\section{Related Work}

%% Contextual Information for Stance Detection in Twitter %%
Due to the informal and colloquial nature of Twitter posts, stance detection on Twitter is a challenging task.
%Further, identifying the stance in individual posts is difficult as they are isolated from their conversational context, such as the topic of discussion or preceding posts which are referred to.
Thus, researchers investigated additional information (i.e., contextual features) which can be helpful to infer the stance, besides the text itself.

% contextualization with external knowledge sources
\paragraph{Additional Information for Stance Detection.}
Previous work employed additional knowledge sources to better inform the model about implicit target-specific information~\citep{du_knowledge2020, sun_knowledge2021, beck-etal-2023-robust}.
%\citet{kawintiranon-singh-2021-knowledge} identify label-relevant tokens and prioritize those during masked language modeling.
\citet{clark-etal-2021-integrating} apply a knowledge infusion method for LLMs by filtering Wikipedia triplets for contextual knowledge.
Similarly, \citet{he-etal-2022-infusing} introduce WS-BERT, which infuses Wikipedia summaries of the target into the stance encoder and reports consistent improvements on target-specific, cross-target, and zero/few-shot stance detection.
\citet{liu-etal-2021-enhancing} uses ConceptNet to identify relevant concept-edge pairs and integrate them via a graph neural network during training.
Going beyond static knowledge bases, \citet{zhang-etal-2024-llm-driven} prompt LLMs to extract the relationship between a tweet and its target as contextual knowledge and inject it into a BART-based classifier, achieving improved results in zero-shot and cross-target stance detection. Additionally, \citet{sucu-etal-2025-exploiting} investigate the impact of enriching LLMs with user profile summaries derived from prior posts, finding that contextual information improves political stance detection performance compared to baseline.
Recently, \citet{nguyen-kim-2025-external} showed that adding Wikipedia or web-search excerpts to LLM prompts degrades stance detection performance by up to 27.9 macro-F1 across eight models, as LLMs tend to align their predictions with the stance of the provided context. 
Our work complements this line of work by studying social-media-intrinsic context rather than encyclopedic knowledge.

% social network metadata
\paragraph{Social Media Metadata for Stance Detection.}
In particular for social networks, several works leveraged the conversational interaction structures to improve the performance of stance detection models~\citep{magdy2016islam, li-etal-2018-structured, porco-goldwasser-2020-predicting, stem2022, sutter-etal-2024-unsupervised}. 
Leveraging Twitter's platform-specific metadata is especially useful for unsupervised user-based stance detection~\citep{thonet2017users, darwish2020, benton-dredze-2018-using, samih-darwish-2021-topical, stem2022}, which aims to identify the stance of a user rather than individual tweets~\citep{10.1145/3404835.3462815}.
For example, \citet{aldayel2019exposed} show that using user's on-topic posts, and a large set of Twitter-specific metadata, such as retweets, replies, likes, followers, and followees, often outperforms a purely content-based approach.

% data augmentation, synthetic data generation
\paragraph{Data Augmentation for Stance Detection.}
Some works explored annotation and generation of additional context information for stance detection.
\citet{jayaram-allaway-2021-human} used human-annotated rationales to improve the reasoning of stance detection models.
In a similar fashion, \citet{math12071119} used Chain-of-Thought (CoT) prompting~\citep{chainofthought2022} to extract explanations for the stance labels which they use subsequently for supervised training.
\citet{gatto-etal-2023-chain} also generate CoT rationales but integrate their embeddings into a transformer-based classifier to mitigate issues like hallucination.
\citet{zhao-etal-2024-zerostance} use ChatGPT to demonstrate the feasibility of cost-efficient dataset generation using LLM prompting without compromising for model generalization using the synthetic data.

% context prompting
\paragraph{Contextualized Prompting for Stance Detection.}
A few works investigated context features for prompting setups directly.
\citet{gambini2024evaluating} investigated the impact of integrating a user's other tweets (i.e., timeline) on the model performance using a framework for user-level zero-shot stance detection.
%They propose a  on Twitter using LLMs, consisting of a topic filtering and agreement detection module.
\citet{Lan2024Agents} employ a multi-agent setup to enhance prompting with linguistic analysis and domain expertise.
\citet{taranukhin-etal-2024-stance} employ a prompting technique which uses in-context learning with examples of CoT reasoning, manually designed and annotated for stance detection.
In addition, they add a task description and use a self-consistency~\citep{wang2023selfconsistency} approach to consolidate multiple generation of output into a majority vote.
Their results show better generalization performance on three different Twitter stance detection benchmarks.

% The gap we address
Unlike prior work on objectively-defined relevance, we systematically evaluate real-world and LLM-generated context for stance detection which is a subjective task where context usefulness is target-dependent.

\begin{figure}[!t]
	\small
	\fbox{%
		\begin{minipage}{0.46\textwidth}
			\colorline{prompt}{You are a stance-labelling assistant.}

			\colorline{prompt}{What stance is expressed towards }\colorline{dataset}{'Face masks' }\colorline{prompt}{ in the tweet below?}

			\colorline{prompt}{Username: }\colorline{sociodemographics}{kaygesmile}

			\colorline{prompt}{User biography: }\colorline{sociodemographics}{We're all in this together but we're not in the same boat \#BLM \#Defendcivilrights \#Antiracist \#Voteblue}

			\colorline{prompt}{Other tweets by user: }
			
			\colorline{sociodemographics}{- RT \@TheRickWilson: Masks are pro-life.}
			
			\colorline{sociodemographics}{- RT \@RexChapman: If these guys can do it, we all can. Please wear your mask around others}

			\colorline{sociodemographics}{- [\dots]}

			\colorline{prompt}{Target description: }\colorline{sociodemographics}{Face masks are protective coverings worn over the nose and mouth, primarily used to reduce the transmission of respiratory infections, including viruses like SARS-CoV-2, which causes COVID-19. [\dots]}

			\colorline{prompt}{Tweet-Specific Context: }\colorline{sociodemographics}{The tweet references the ongoing debates surrounding face masks during the COVID-19 pandemic and gun rights in the United States, particularly the Second Amendment, which protects the right to keep and bear arms. The mention of people refusing to wear masks highlights the resistance some individuals showed towards public health measures designed to mitigate the spread of the virus, often citing personal freedom. [\dots]}

			\colorline{prompt}{Tweet: }\colorline{dataset}{Is it lost on anyone that the exact same people who moan and refuse to wear a mask to save another human life are the EXACT SAME people who doggedly defend the 2nd amendment after 50 people get shot dead by an AK-47? \#therightiswrong \#COVID-19}

			\colorline{prompt}{Stance }\colorline{dataset}{(favor, against, none):}
		\end{minipage}
	}
	\caption{Prompt with additional contextual features to predict the stance of a Twitter post. The different parts of the prompt are highlighted, i.e. \colorline{prompt}{instruction}, \colorline{sociodemographics}{contextual information} and \colorline{dataset}{dataset input}. Example drawn from the covid19-glandt dataset~\citep{glandt-etal-2021-stance}.}
	\label{ch:prompting_noisy_domains:fig:sd-prompt-example}
\end{figure}

\section{Experimental Setup}

\subsection{Prompting}

% describe the prompt 
We employ a standard prompting setup for stance detection by providing a short task instruction including the answer possibilities (i.e., labels), the tweet text, and the target.
An example containing all possible contextual features for a single prompt is given in \Cref{ch:prompting_noisy_domains:fig:sd-prompt-example}.
We conduct zero-shot and few-shot experiments and additionally compare with CoT prompting~\citep{chainofthought2022}.
Further, we employ an LLM-as-a-Judge prompting approach following the method introduced by~\citep{DBLP:conf/nips/ZhengC00WZL0LXZ23}, in which the same model used for the initial classification is prompted to re-evaluate the prediction and revise the label if the original classification is deemed inadequate.

% obtaining predictions using structured outputs
To obtain the predictions for the stance classification setting, we employ vLLM's \textit{structured outputs}\footnote{\url{https://docs.vllm.ai/en/latest/features/structured_outputs/}} configuration. 
We used default parameters, except temperature which was set to $0.5$.
Performance results are averaged across three different seeds to account for non-determinism in LLM inference and provide more reliable performance estimates.
We provide the exact prompts and details about the prompting configuration in the Appendix~\ref{app:sec:models}.

\subsection{Datasets}
\label{ssec:datasets}

\begin{table*}[t]
	\centering
	\resizebox{1.00\textwidth}{!}{%
		\begin{tabular}{p{3.5cm}p{1.5cm}p{2cm}p{1cm}p{7cm}}
			\toprule
			\textbf{Dataset (Lang)} & \textbf{Size} & \textbf{Filtered Size} & \textbf{Topics} & \textbf{Labels} \\
			\midrule
			covid19-de (DE) & 3,750   & 3,046 (-19\%) & 1 & none (80\%), pro (10\%), con (9\%) \\
            covid19-glandt (EN) & 6,133   & 1,017 (-83\%) & 4 & favor (42\%), none (34\%), against (22\%)\\
			semeval2016t6 (EN) & 4,163   & 859 (-79\%) & 6 & against (44\%), favor (28\%), none (26\%) \\
			wtwt (EN) & 51,284  & 9,263 (-82\%) & 5 & comment (39\%), unrelated (38\%), support (12\%), refute (9\%) \\
			\bottomrule
		\end{tabular}
	}
	\caption{Twitter Stance Detection datasets and their characteristics. Metadata could not be retrieved for a considerable portion of tweets, resulting in a strong reduction of instances after filtering.}
	\label{ch:prompting_noisy_domains:tab:twitter_stance_datasets}
\end{table*}

In the following, we provide a brief overview of each dataset and an overview of the statistics of all datasets in \Cref{ch:prompting_noisy_domains:tab:twitter_stance_datasets}.

\paragraph{covid19-glandt, semeval2016t6, wtwt.} We use three existing and heterogeneous Twitter stance detection datasets in our experiments.
\texttt{semeval2016t6}~\citep{mohammad-etal-2016-semeval} consists of tweets discussing either a person, organization, or a topic of public interest (e.g., Hillary Clinton, Climate Change).
The \texttt{covid19-glandt}~\citep{glandt-etal-2021-stance} dataset covers topics prominent in U.S.-centric debates during Covid-19 period, such as \textit{Stay at home orders} and \textit{Dr. Anthony Fauci}.
The \texttt{wtwt} ("Will-They-Won't-They") by \citet{conforti-etal-2020-will} is a Twitter stance detection in the context of mergers and acquisitions (M\&A), annotated by financial domain experts.
It focuses on five M\&A operations involving major companies from healthcare and entertainment industries.
Where necessary, we converted topic labels to readable format (e.g., \textit{face\_masks} to \textit{Face masks}) as described in \Cref{app:sec:datacleaning}.

\paragraph{covid19-de.} Users with different cultural backgrounds behave differently online~\citep{trepte2016socialmediaculture}. 
Thus, we extend our analysis to another language and introduce \texttt{covid19-de}, a new dataset consisting of tweets in German discussing governmental measures to contain the spread of Covid-19.
The crawling period ranges from 01.01.2020 until 31.03.2022.
After pre-processing, the dataset consisted of 23,972,587 tweets.
We leveraged existing annotation guidelines~\citep{mohammad-etal-2016-semeval, beck-etal-2021-investigating} and recruited seven German-speaking student annotators from a political communication science course. 
We trained the annotators in multiple rounds until substantial agreement was achieved (Krippendorff $\alpha>0.67$), on a set of tweets (n=250) annotated by a political science communication expert.
Finally, we sampled 3,500 tweets equally across the full time range and assigned each annotator 500 tweets.

Institutional actors may amplify certain positions, generate media attention, or shape public discourse without representing the distribution of opinion among the general population. 
To account for that we classified all accounts into four \textit{actor classes} (i.e., politician, journalist, scientist, citizen) using a semi-automatic classification procedure based on the Twitter handle and user biography.
Further, we assigned each non-citizen account a fine-grained \textit{author group} based on information given in their user biography (e.g., assigning \textit{Die Grünen} due to a reference to the green party in Germany).
More details can be found in the Appendix~\ref{app:sec:datacollection}.

\subsection{Contextual Features}
\label{ch:finalch:sec:contextual_info}

\subsubsection{Twitter Metadata}
% author's username, profile description, and other tweets by the same user.

For each dataset, we retrieved additional metadata using the official Twitter API\footnote{\url{https://developer.x.com/}} and filtered the datasets to include only tweets for which we could retrieve metadata (see \Cref{ch:prompting_noisy_domains:tab:twitter_stance_datasets}).
This led to a large reduction of dataset size (~80\%) due to tweets being deleted, or accounts being suspended or set to private mode.
However, this did not affect the representativeness of the datasets in terms of class imbalance (see for details).
Due to size limitations, we provide examples of retrieved metadata in \Cref{app:sec:twittermetadata}.
Unfortunately, we were not able to retrieve other tweets by the same users for the covid19-de dataset, as during the crawling period our Twitter Academic API access was revoked.
%We note that we performed crawling of all metadata using an academic access for retrieving historical data, before Twitter's policy change in March 2023.} to retrieve the beforementioned additional metadata.
%We then filtered the datasets to include only tweets for which we could retrieve metadata.
In the following, we describe each metadata field we retrieve.

\paragraph{Username.} Users can dynamically adjust their usernames on Twitter. 
Prominent individuals, due to their frequent public appearances and media coverage, often make their stances more readily inferable from their usernames by adapting them in reaction to current events (e.g., adding a flag to demonstrate solidarity with a nation). 
Additionally, many users adopt made-up names that can subtly reflect their political affiliations or personal beliefs (e.g., \textit{RealPatriot1776}). 
However, a significant number of usernames do not provide clear indications of stance.
This allows to evaluate whether a model can effectively differentiate between usernames that are informative and those that are neutral or ambiguous.

\paragraph{User Biography.} We retrieve the self-written user biography, as users frequently provide personal information that signals their beliefs or affiliations. 
For instance, when users describe themselves as members of specific political parties, it serves as a strong indicator of their alignment with that party's stance. 

\paragraph{User Tweets.} Lastly, previous research~\citep{burfoot-etal-2011-collective, samih-darwish-2021-topical, gambini2024evaluating} has demonstrated that analyzing a user's tweet history can be highly informative in understanding their position towards topics of discussions. 
Therefore, we retrieve other tweets by the same user as the author of the text we aim to conduct stance classification on.
To limit the number of tweets we use for our experiments, we compute the cosine similarities between all tweets by the same user and select the ten most similar ones to the target tweet.
We use SentenceBERT~\citep{reimers-gurevych-2019-sentence} for computing the cosine similarities. 
Specifically, we use the \texttt{all-mpnet-base-v2} model.
Due to policy changes during our crawling period which have significantly restricted free access to Twitter data for research purposes, we were not able to retrieve other user tweets for the \texttt{covid19-de} dataset.

%\paragraph{Measure discussed.} We asked annotators for the specific Covid19-related measure being discussed in the tweet. 
%Annotators could choose one or multiple of eight broad categories: (1) unspecified general measures, (2) shutdown and lockdown, (3) general prevention measures such as wearing masks, (4) person count limits and closures, (5) prohibitions such as border closure, (6) vaccination, (7) government investments such as compensations, or (8) other measure.

\subsubsection{LLM-Generated Context}

To provide additional context information for the tweets, we generate synthetic data using an LLM.
Specifically, we generate \textit{Target Description} and \textit{Tweet-Specific Context}, using the prompt templates provided in \Cref{ch:prompting_noisy_domains:tab:prompts}.
We use GPT-4\footnote{The model ID in use is \texttt{gpt-4o-mini-2024-07-18}}~\citep{openai2024gpt4} to generate the context information for all tweets and targets, using the official OpenAI API\footnote{\url{https://openai.com/api/}}.
We set \texttt{max\_completion\_tokens=200} and otherwise use default parameters, i.e., \texttt{top\_p=1}, \texttt{temperature=1.0}.
%\Cref{ch:prompting_noisy_domains:tab:twitter_metadata} provides and abbreviated example of the generated context information.
The exact prompts and examples are provided in \Cref{app:sec:twittermetadata}.

\paragraph{Target Description.} Target information has been shown to be an informative feature for stance detection in previous work~\citep{stab-etal-2018-cross, reimers-etal-2019-classification}.
However, the target name itself often does not carry enough information to infer the stance of the tweet.
Thus, we instruct the LLM to generate a short description of the target.

\paragraph{Tweet-Specific Context.}

The target itself often has multiple aspects of discussion which are potentially referenced in the tweet but are not apparent from the target itself.
For example, tweets about \textit{Hillary Clinton} from the \texttt{semeval2016t6} dataset~\citep{mohammad-etal-2016-semeval} mention the 2012 Bengazhi attack during Hillary Clinton's position as secretary of state.
To extract these aspects, we instruct the model to generate a short description of the target in the context of the tweet.

\subsection{Models}

%Instruction-tuned LLMs have been shown to be effective for text classification tasks, due to the large coverage of tasks in the annotated data used for instruction tuning.
%We leverage different reasoning and instruction-tuned models with various parameter sizes from the \texttt{OPT-IML}~\citep{iyer2023optiml}, \texttt{Mistral-Instruct}~\citep{jiang2023mistral7b} and \texttt{Llama3.1}~\citep{dubey2024llama3} model families. 
We leverage different reasoning and instruction-tuned models with a mid-sized number of parameters; we use \texttt{GPT-OSS-20B} \citep{openai2025gptoss120bgptoss20bmodel}, \texttt{Qwen2.5-32B} \citep{qwen2025qwen25technicalreport}, \texttt{Gemma3-27B} \citep{gemmateam2025gemma3technicalreport}, and \texttt{Ministral3-14B} \citep{liu2026ministral3}.
%More details about the models are provided in \Cref{app:sec:models}.
%For \texttt{OPT-IML}, we use the 1.3B, and 70B parameter versions.
%These have been fine-tuned on a large collection of instruction-style tasks.

%For the multi-lingual \texttt{Mistral-Instruct}, we use the 7.3B parameter version as well as the instruction-tuned versions (12.2B, 22.2B) of the more recently released version (i.e., NeMo) with larger context length and an improved tokenizer\footnote{For more details, please refer to the accompanying blog post at \url{https://mistral.ai/news/mistral-nemo/}}.

%The multi-lingual \texttt{Llama3.1} models were fine-tuned using post-training procedures based on supervised finetuning, rejection sampling, and direct preference optimization~\citep{rafailov2023dpo}.
%We use the 8B, and 70B parameter versions in our experiments.

%\Cref{ch:prompting_noisy_domains:tab:instruction_tuned_models} provides an overview of the models used in our experiments.

\section{Results}

\begin{table*}[t!]
    \small
    \centering                                                                         \resizebox{.99\textwidth}{!}{%
    \begin{tabular}{lcccc|cccc}
        \toprule
        & \multicolumn{4}{c|}{Zero-Shot} & \multicolumn{4}{c}{Few-Shot (n=10)} \\
      \cmidrule(lr){2-5} \cmidrule(lr){6-9} Context & covid19-de & covid19-glandt & semeval2016t6 & wtwt & covid19-de & covid19-glandt & semeval2016t6 & wtwt \\     \midrule                                                                         Username & $-2.75_{\pm 4.60}$ & $\mathbf{+2.80}_{\pm 2.49}$ & $+0.19_{\pm 3.14}$ & $-2.87_{\pm 3.16}$ & $-1.32_{\pm 2.92}$ & $+1.09_{\pm 1.84}$ & $+1.06_{\pm 0.92}$ & $+0.95_{\pm 
  3.88}$ \\                                                                                User Description & $-1.43_{\pm 2.96}$ & $\mathbf{+2.73}_{\pm 1.86}$ & $+1.45_{\pm 2.77}$ & $-4.02_{\pm 3.55}$ & $-0.19_{\pm 1.90}$ & $+1.53_{\pm 2.11}$ & $+0.36_{\pm 1.82}$ &   
  $-5.61_{\pm 2.62}$ \\                                                                    Other Tweets & -- & $\mathbf{+4.45}_{\pm 2.93}$ & $-0.03_{\pm 8.38}$ & $-3.20_{\pm 3.46}$ & -- & $+0.79_{\pm 6.00}$ & $-5.17_{\pm 2.57}$ & $-6.47_{\pm 4.40}$ \\                  Target Description & $-0.35_{\pm 2.39}$ & $\mathbf{+3.52}_{\pm 1.71}$ & $\mathbf{+5.10}_{\pm 1.42}$ & $\mathbf{+3.51}_{\pm 2.86}$ & $+1.93_{\pm 2.12}$ & $+2.94_{\pm 3.24}$ & $\mathbf{+3.95}_{\pm 1.42}$ &   
  $+0.20_{\pm 5.94}$ \\                                                                     Tweet-Specific Context & $-4.12_{\pm 0.59}$ & $-0.95_{\pm 5.56}$ & $-2.18_{\pm 9.02}$ & $-3.61_{\pm 4.87}$ & $-3.32_{\pm 1.52}$ & $-0.43_{\pm 7.22}$ & $-2.58_{\pm 4.63}$ 
  & $-5.40_{\pm 5.70}$ \\                                                                   All & -- & $+4.67_{\pm 6.39}$ & $-2.66_{\pm 8.37}$ & $-2.36_{\pm 5.93}$ & -- & $+3.36_{\pm 8.40}$ & $-3.51_{\pm 5.71}$ & $-2.82_{\pm 17.81}$ \\
  \bottomrule
  \end{tabular}}\caption{Mean $\Delta$F1 (in percentage points) across four reasoning models (\texttt{GPT-OSS-20B}, \texttt{Qwen2.5-32B}, \texttt{Gemma3-27B}, \texttt{Ministral3-14B}) when adding each context feature,     
  relative to the no-context baseline of the same reasoning setting. Across-model standard deviation in subscripts. Results in bold signal clear improvement above standard deviation. Cells marked `-' have no runs due to missing data (see \cref{ch:finalch:sec:contextual_info}).}
      \label{tab:meta_delta_f1}
\end{table*} 

\subsection{Contextual Feature Importance}

We provide the mean difference in macro-averaged F1 ($\Delta$F1) across the four models when adding each context feature in \Cref{tab:meta_delta_f1}.
Context features are beneficial in both settings, but the effect is consistently dampened in few-shot compared to zero-shot.

\textbf{Target Description:} Across all features, the LLM-generated \textit{Target Description} is the best generalizing feature with an improvement in almost all settings except zero-shot \texttt{covid19-de} and with low standard deviation (std$\leq3.3$pp).
Most likely, as the target is often not or only implicitly mentioned in the text~\citep{augenstein-etal-2016-stance}, adding more target information helps in resolving the implicit reference to the target in the text.
This is in line with previous work stating that providing target information generally improves robustness of stance detection models across different benchmarks~\citep{schiller2021benchmark, beck-etal-2023-robust}.
% Interpretation?
\textbf{Tweet-Specific Context:}
Counterintuitively and despite being the more targeted, LLM-generated variant, \textit{Tweet-Specific Context} has a negative effect in every setting across datasets.
We analyzed the cases in which adding \textit{Other Tweets} changes a \textit{correct} baseline prediction
(see \Cref{app:sec:tweetspecificcontext_flips}).
The resulting incorrect predictions are heavily concentrated on non-neutral stance classes indicating a bias induced by the added context.
Using a rule-based lexicon approach, we map the generated context into two subgroups, based on whether it contains stance-bearing words.
Analyzing the performance changes stratified by these two subgroups, we find that the performance on the three person-centric datasets (\texttt{covid19-de}, \texttt{covid19-glandt}, \texttt{semeval2016t6}) suffers from stance-bearing context in the LLM-generated context.
This highlights the challenge to provide context without any semantic interpretation using LLMs.

\textbf{Other Tweets:} In contrast to previous work on \textit{supervised} stance detection~\citep{aldayel2019exposed, gambini2024evaluating}, providing \textit{Other Tweets} by the same user is only beneficial for \texttt{covid19-glandt}. 
As we use up to ten different tweets which can substantially vary in length, we hypothesize that the increasing length of the input prompt is more likely to ''distract'' the model.
Similar observations have been discussed in related work~\citep{min-etal-2022-rethinking, liu-etal-2024-lost}, especially if the additional context is semantically related to the text of interest~\citep{wu2024}.
However, stratifying $\Delta$F1 by the per-instance length of the \textit{Other Tweets} context (see \Cref{app:sec:other_tweets_length}) does not show a monotonic decline.  
On \texttt{covid19-glandt} zero-shot the gain is largest in both the shortest \emph{and} longest length quartile 
%($+4.81$pp and $+6.06$pp)
, and on \texttt{wtwt} few-shot the regression is roughly constant across length quartiles. %($-6.6$ to $-7.1$pp).
Thus, we investigated the cases in which adding \textit{Other Tweets} changes a \textit{correct} baseline prediction (\Cref{app:sec:othertweets_flips}). 
As for \textit{Tweet-Specific Context}, the resulting incorrect predictions are heavily concentrated on stance-bearing classes (e.g.\ more than 90\% \emph{against} or \emph{favor} on \texttt{covid19-glandt}), while the neutral \emph{none} class is rarely changed.
Adding the user's recent tweets make the model project an opinionated stance even when the correct label is neutral.

\textbf{All:} Having a more complete context by providing \textit{All} features is only beneficial  for \texttt{covid19-glandt} as a direct result of almost all individual context features improving upon their baselines without any context. 
%The std of 17.81pp on wtwt few-shot is driven by Ministral's outlier +22.93 — flag this so the table doesn't mislead a skim-reader.

\begin{table*}[t!]
    \small
    \centering
    \resizebox{.99\textwidth}{!}{%
    \begin{tabular}{c p{3cm}p{2.5cm}p{2.5cm}p{2.5cm}p{2.5cm}}
    \toprule
 & Setting & covid19-de & covid19-glandt & semeval2016t6 & wtwt \\
    \midrule
    \multirow{4}{*}{\rotatebox{90}{No context}} & Zero-Shot & $45.23_{\pm 1.33}$ & $65.10_{\pm 1.39}$ & $62.45_{\pm 1.39}$ & $60.29_{\pm 3.50}$ \\
     & Few-Shot (n=10) & $48.60_{\pm 1.13}$ & $64.06_{\pm 1.42}$ & $63.43_{\pm 0.61}$ & $\mathbf{67.28}_{\pm 1.26}$ \\
     & CoT & $45.32_{\pm 1.01}$ & $55.32_{\pm 1.19}$ & $53.51_{\pm 0.64}$ & $58.84_{\pm 2.17}$ \\
     & LLM-as-a-Judge & $\mathbf{51.58}_{\pm 0.79}$ & $65.22_{\pm 2.06}$ & $64.53_{\pm 1.48}$ & $60.22_{\pm 1.01}$ \\
    \midrule
    \multirow{6}{*}{\rotatebox{90}{\shortstack{Zero-Shot \\ + Context}}} & Username & $44.85_{\pm 0.66}$ & $70.99_{\pm 1.50}$$^{*\dag}$ & $67.15_{\pm 0.36}$$^{*\dag}$ & $57.54_{\pm 1.97}$ \\
     & User Description & $45.73_{\pm 1.04}$ & $69.63_{\pm 1.41}$$^{*\dag}$ & $66.71_{\pm 0.52}$$^{*\dag}$ & $56.49_{\pm 1.18}$ \\
     & Other Tweets & - & $67.47_{\pm 1.37}$ & $65.36_{\pm 0.79}$$^{*}$ & $56.15_{\pm 0.59}$ \\
     & Target Description & $47.66_{\pm 1.01}$$^{*}$ & $70.82_{\pm 0.96}$$^{*\dag}$ & $67.21_{\pm 1.25}$$^{*}$ & $59.68_{\pm 0.92}$ \\
     & Tweet-Specific Context & $41.23_{\pm 1.14}$ & $68.42_{\pm 0.70}$$^{*\dag}$ & $63.82_{\pm 0.45}$ & $54.02_{\pm 1.38}$ \\
     & All & - & $71.10_{\pm 0.49}$$^{*\dag}$ & $63.36_{\pm 0.71}$ & $52.60_{\pm 0.91}$ \\
    \midrule
    \multirow{6}{*}{\rotatebox{90}{\shortstack{Few-Shot \\ + Context}}} & Username & $49.25_{\pm 0.76}$ & $64.88_{\pm 4.19}$ & $64.37_{\pm 2.24}$ & $66.88_{\pm 0.57}$ \\
     & User Description & $50.75_{\pm 0.67}$$^{*}$ & $68.03_{\pm 2.87}$$^{*}$ & $66.45_{\pm 1.31}$$^{*}$ & $64.93_{\pm 0.93}$ \\
     & Other Tweets & - & $57.89_{\pm 0.83}$ & $61.13_{\pm 3.03}$ & $60.13_{\pm 1.66}$ \\
     & Target Description & $50.80_{\pm 0.53}$$^{*}$ & $70.77_{\pm 1.84}$$^{*\dag}$ & $\mathbf{68.68}_{\pm 0.76}$$^{*\dag}$ & $65.49_{\pm 0.78}$ \\
     & Tweet-Specific Context & $46.22_{\pm 1.31}$ & $71.54_{\pm 0.49}$$^{*\dag}$ & $65.81_{\pm 2.04}$ & $64.79_{\pm 1.23}$ \\
     & All & - & $\mathbf{72.48}_{\pm 2.05}$$^{*\dag}$ & $65.97_{\pm 1.35}$ & $62.57_{\pm 0.52}$ \\
    \bottomrule
    \end{tabular}
    }
    \caption{Macro F1 results for \texttt{GPT-OSS-20B}. Bold results are best per dataset, $^{*}$ is for stat. significant better performance than the corresponding reasoning-type baseline (i.e., zero-shot + context vs. no-context zero-shot and few-shot + context vs no-context few-shot), and $^{\dag}$ is for stat. significant better performance than \textit{all} reasoning-type baselines (i.e., no-context zero-shot, no-context few-shot, CoT, LLM-as-a-Judge) using a paired bootstrap test ($p < 0.05$). Cells marked `-' have no runs due to missing data (see \cref{ch:finalch:sec:contextual_info}).}
    \label{tab:gptoss20b_results_extended}
\end{table*}

\subsection{Model-specific Results}

We provide the detailed model results for \texttt{GPT-OSS-20B} (\Cref{tab:gptoss20b_results_extended}) which has the strongest average performance across datasets and context features. 
As highlighted previously, \textit{Target Description} is the most reliable feature with the strongest gains on \texttt{covid19-glandt} (+5.72pp zero-shot, +6.71pp few-shot) and \texttt{semeval2016t6} (+4.76pp zero-shot, +5.25pp few-shot).
Notably, the gain holds in both zero-shot and few-shot settings.
This confirms that the value of explicit target information is not subsumed by the in-context demonstrations.
On \texttt{wtwt}, by contrast, neither setting yields a significant improvement.
We attribute this to the highly specific nature of the M\&A target pairs for which a short LLM-generated description provides little additional disambiguation.

% Other Tweets — model-specific instantiation                                         
For \textit{Other Tweets}, the per-model result provide a more fine-grained view in comparison to the cross-model results in \Cref{tab:meta_delta_f1}.
On \texttt{GPT-OSS-20B}, the only significant gain is $+2.91$pp on \texttt{semeval2016t6} in the zero-shot setting, while in the few-shot setting the feature regresses by $6.17$pp on \texttt{covid19-glandt} and $7.15$pp on \texttt{wtwt}.
                  
% CoT and LLM-as-a-Judge baselines
\textbf{Baselines:} We additionally display the performance of other prompting baselines, i.e. CoT prompting and LLM-as-a-Judge.
Surprisingly, \textit{CoT} prompting regresses below the zero-shot baseline on three of four datasets ($-9.78$pp on \texttt{covid19-glandt}, $-8.94$pp on \texttt{semeval2016t6}, $-1.45$pp on \texttt{wtwt}, with \texttt{covid19-de} essentially unchanged at $+0.09$pp), suggesting that explicit reasoning chains do not benefit short-text stance classification when no additional context is provided. 
The regression of \textit{CoT} below the zero-shot baseline holds for all four models on \texttt{covid19-glandt}, \texttt{semeval2016t6}, and \texttt{wtwt}, but the magnitude varies considerably (see  Appendix~\ref{app:sec:results} for detailed results). 
%\texttt{GPT-OSS-20B} shows moderate drops (up to $-9.78$pp), \texttt{Ministral-14B} regresses by 10--30pp, and \texttt{Qwen2.5-32B} and \texttt{Gemma3-27B} regress by 18--49pp.
%The only \textit{CoT} cell with a positive effect across all four models is \texttt{Gemma3-27B} on \texttt{covid19-de} ($+3.81$pp).
%We leave a fuller interpretation of this single-cell exception to future work. 
\textit{LLM-as-a-Judge}, in contrast, is competitive with the zero-shot baseline and yields the strongest overall result on \texttt{covid19-de} (51.58), exceeding all context-augmented configurations on this dataset.
Our analysis shows that the gains on this dataset are due to the method's tendency to correct predictions into a neutral stance (see \Cref{app:judge_analysis}).
Due to the label imbalance on \texttt{covid19-de}, this correction is particularly helpful. 
On the other more balanced datasets, the improvements are on-par with the zero-shot prompting baselines due to prediction correction leading to both positive and negative changes.

\subsection{Instance-level Analysis}

\begin{table}[h!]
\centering
\resizebox{.49\textwidth}{!}{%
\begin{tabular}{lllll}
\toprule
Feature & covid19-de & covid19-glandt & semeval2016t6 & wtwt \\
\midrule
\multirow{3}{3cm}{\textbf{Target Description}}& {\color{green}$\uparrow$} 9\% (289) & {\color{green}$\uparrow$} 12\% (125) & {\color{green}$\uparrow$} 13\% (114) & {\color{green}$\uparrow$} 8\% (738) \\
& {\color{red}$\downarrow$} 6\% (175) & {\color{red}$\downarrow$} 6\% (65) & {\color{red}$\downarrow$} 8\% (73) & {\color{red}$\downarrow$} 8\% (746) \\
& {\color{gray}--} 85\% (2583) & {\color{gray}--} 81\% (827) & {\color{gray}--} 78\% (672) & {\color{gray}--} 84\% (7780) \\
\midrule
\multirow{3}{3cm}{\textbf{Username}}& {\color{green}$\uparrow$} 7\% (210) & {\color{green}$\uparrow$} 12\% (117) & {\color{green}$\uparrow$} 12\% (107) &                 
{\color{green}$\uparrow$} 7\% (619) \\
& {\color{red}$\downarrow$} 7\% (223) & {\color{red}$\downarrow$} 6\% (56) & {\color{red}$\downarrow$} 8\% (68) & {\color{red}$\downarrow$} 8\% (723) \\
& {\color{gray}--} 86\% (2612) & {\color{gray}--} 83\% (844) & {\color{gray}--} 80\% (684) & {\color{gray}--} 86\% (7921) \\
\midrule
\multirow{3}{3cm}{\textbf{Tweet-Specific Context}}& {\color{green}$\uparrow$} 6\% (193) & {\color{green}$\uparrow$} 15\% (150) & {\color{green}$\uparrow$} 17\% (147) & {\color{green}$\uparrow$} 7\% (680) \\
& {\color{red}$\downarrow$} 13\% (410) & {\color{red}$\downarrow$} 11\% (111) & {\color{red}$\downarrow$} 14\% (120) & {\color{red}$\downarrow$} 14\% (1321) \\
& {\color{gray}--} 80\% (2443) & {\color{gray}--} 74\% (755) & {\color{gray}--} 69\% (591) & {\color{gray}--} 78\% (7262) \\
\midrule                                                               \multirow{3}{3cm}{\textbf{User Description}}& {\color{green}$\uparrow$} 8\% (230) & {\color{green}$\uparrow$} 12\% (125) & {\color{green}$\uparrow$} 14\% (118) & {\color{green}$\uparrow$} 7\% (650) \\
& {\color{red}$\downarrow$} 7\% (222) & {\color{red}$\downarrow$} 8\% (79) & {\color{red}$\downarrow$} 10\% (82) & {\color{red}$\downarrow$} 9\% (824) \\
& {\color{gray}--} 85\% (2594) & {\color{gray}--} 80\% (813) & {\color{gray}--} 77\% (659) & {\color{gray}--} 84\% (7790) \\
\midrule                                                               \multirow{3}{3cm}{\textbf{Other Tweets}}& -- & {\color{green}$\uparrow$} 14\% (143) & {\color{green}$\uparrow$} 16\% (136) & {\color{green}$\uparrow$} 8\% (717) \\
& -- & {\color{red}$\downarrow$} 11\% (114) & {\color{red}$\downarrow$} 12\% (104) & {\color{red}$\downarrow$} 11\% (1014) \\
& -- & {\color{gray}--} 75\% (759) & {\color{gray}--} 72\% (619) & {\color{gray}--} 81\% (7532) \\
\bottomrule                                                            \end{tabular}}        
\caption{Instance-level impact of context integration for \texttt{GPT-OSS-20B} in the zero-shot setting, compared with the no-context baseline. Within each cell, the first row shows instances where the prediction is improved (baseline wrong, context correct), the second row instances where it is degraded (baseline correct, context wrong), and the third row instances where the correctness is unchanged. Counts are averaged across three seeds and rounded to the nearest integer. Cells marked `-' have no runs due to missing data (see \cref{ch:finalch:sec:contextual_info})}
\label{ch:finalch:tab:instance_level_comparison_gptoss20b_zeroshot}
\end{table} 

To better understand the influence of features on instance-level predictions, we conduct an analysis of the predictions made by \texttt{GPT-OSS-20B}.
Results are presented in \Cref{ch:finalch:tab:instance_level_comparison_gptoss20b_zeroshot}.

In general, we observe that all contextual features improve a non-negligible number of instances on each dataset, with improvement rates between 6\% and 17\%.
This confirms our initial hypothesis that all context features potentially provide valuable information for stance detection.
However, at the same time the model predictions are influenced negatively, with correct baseline predictions changed due to the integration in 5--19\% of cases.
This hints at the model's lack of ability to effectively differentiate between informative and misleading contextual information.
The main difference between favourable (i.e., \textit{Target Description}) and unfavourable context features (i.e., \textit{Tweet-Specific Context} and \textit{Other Tweets}) is the ratio between improved and negatively influenced instances.
This is in alignment with the aggregated results (see \Cref{tab:meta_delta_f1}). 
\textit{Tweet-Specific Context} stands out as a high-variance feature. 
It produces among the highest improvement rates (up to 17\%) but also the highest degradation rates (up to 14\%) in the zero-shot setting.
Across zero-shot and few-shot settings the per-feature rankings are stable. 
The one substantive change is that \textit{Other Tweets} degrades 5--7pp more often in few-shot than in zero-shot, while improving comparably few additional instances (\Cref{app:ch:finalch:tab:instance_level_comparison_gptoss20b_fewshot}).

\subsection{covid19-de Context}

\begin{table}[t]
    \centering
    \resizebox{.49\textwidth}{!}{%
    \begin{tabular}{llcccc}
    \toprule
    Setting & Feat. & \shortstack{Citizen\\\scriptsize $n{=}2321$} & \shortstack{Media\\\scriptsize $n{=}424$} & \shortstack{Politician\\\scriptsize $n{=}240$} & \shortstack{Scientist\\\scriptsize $n{=}61$} \\
    \midrule
    \multirow{2}{*}{ZS} & AC & {\color{red}$\downarrow$} -0.97 & {\color{red}$\downarrow$} -2.03 & {\color{green}$\uparrow$} +0.87 & {\color{red}$\downarrow$} -1.44 \\
     & AG & {\color{red}$\downarrow$} -1.29 & {\color{red}$\downarrow$} -2.46 & {\color{red}$\downarrow$} -0.36 & {\color{green}$\uparrow$} +0.25 \\
    \midrule
    \multirow{2}{*}{FS} & AC & {\color{green}$\uparrow$} +0.60 & {\color{red}$\downarrow$} -0.64 & {\color{red}$\downarrow$} -2.15 & {\color{green}$\uparrow$} +0.07 \\
     & AG & {\color{red}$\downarrow$} -0.23 & {\color{red}$\downarrow$} -2.72 & {\color{red}$\downarrow$} -0.86 & {\color{red}$\downarrow$} -2.02 \\
    \bottomrule
    \end{tabular}
    }
    \caption{Per-class macro-F1 change ($\Delta$, in pp) over the no-context baseline when injecting author annotations for \texttt{GPT-OSS-20B} on \texttt{covid19-de}. Settings: zero-shot (ZS), few-shot (FS); features: author class (AC), author group (AG). Author classes are partitioned by the dataset's \texttt{class} field; $n$ denotes the number of test instances per class.}
    \label{ch:finalch:tab:semisupervised_perclass_gptoss20b}
\end{table}

Beyond the LLM-generated context features, we further investigate whether author metadata, obtained via semi-automatic assignment (\Cref{ssec:datasets}) on \texttt{covid19-de}, provides beneficial information for stance detection. 
We evaluate the coarse four-way \textit{Author Class} (citizen, media, political, scientist) and fine-grained \textit{Author Group} attribute as context feature. 
Results for \texttt{GPT-OSS-20B} are displayed in \Cref{ch:finalch:tab:semisupervised_perclass_gptoss20b}.
Neither annotation significantly outperforms the no-context baseline.
This finding is consistent across models and context features (see \Cref{ch:finalch:tab:semisupervised_overall_delta} in the appendix).
Stratifying $\Delta$F1 by author class for \texttt{GPT-OSS-20B} shows the regression is most pronounced for \textit{media} and \textit{political} authors, where an improvement via the affiliation signal could be expected.
We conclude that the additional annotation effort does not translate into stance-detection gains for LLMs in the zero-shot or few-shot setting.

\section{Conclusion}

Stance detection on Twitter is challenging due to noisy, brief, context-dependent language. While prior work shows that metadata improves \textit{supervised} stance detection, it is unclear whether LLMs can leverage such context, given their sensitivity to prompt design.
We address this gap by evaluating contextual features from tweet metadata and LLM generation across four Twitter benchmarks and multiple reasoning LLMs.

Notably, using LLM-generated target descriptions boosts performance across all datasets, while for other contextual data, the improvements on some instances are offset by corresponding declines on others.
Our results highlight the potential of context integration in noisy domains but expose limitations in the robustness of prompt-based approaches.
Overall, our findings motivate further research in differentiating useful from noisy context in prompting setups and extend the existing line of research on contextual subjectivity~\citep{wu-etal-2024-instructing, wu2024} to subjective tasks like stance detection, where relevant context cannot always be objectively defined.

\section*{Limitations}

\paragraph{Prompt Design}

We conduct a large number of experiments.
Therefore, we did not aim to identify the best-performing prompt for each model and dataset. 
Prior work has shown that model performance is highly dependent on the choice of prompt~\citep{webson-pavlick-2022-prompt, min-etal-2022-rethinking, lu-etal-2022-fantastically}.
Thus, prompt design remains a critical factor to consider when using LLMs, and future work should focus on optimizing prompts to ensure the best possible performance.

\paragraph{Limited Access to Twitter Data}

One of the primary limitations of this study stems from Twitter's recent policy changes, which have significantly restricted free access to Twitter data for research purposes. 
This limitation hinders the free-of-charge retrieval of valuable contextual information, such as user biographies and interaction histories, for existing datasets.
It affected our experiments in a way that we could not finish the retrieval of all metadata for the covid19-de dataset.
While Twitter was the primary source of research data for many social media studies in the past, the current restrictions have made it challenging to conduct research in this area.
Future projects must take these restrictions into account.

\section*{Ethics Statement}

In this work, we integrate contextual user information as auxiliary signals to improve the performance of stance detection models. 
This raises significant ethical considerations relating to privacy, informed consent, potential misuse, and algorithmic fairness. 
We elaborate on these concerns below and outline the mitigation strategies adopted in this study.

\paragraph{Justification of Contextual User Data Usage} Stance detection is a task inherently sensitive to subjectivity, ambiguity, and context-dependence. 
Prior research in computational social science and political communication has shown that individual-level contextual signals can meaningfully shape or clarify the interpretation of isolated statements, especially in online discourse environments such as Twitter, where posts are short and potentially ambiguous. 
Incorporating metadata can enhance a model’s ability to disambiguate stance by aligning the current statement with broader patterns in user behavior. 
%This follows theoretical work on audience design, identity performance, and homophily in digital communication, which justifies the modeling of users as latent variables in stance or opinion analysis tasks (cf. Volkova et al., 2017; Preoţiuc-Pietro et al., 2017).

From a technical standpoint, our use of user context aims to augment noisy and under-specified signals with richer, weakly supervised features. 
This approach aligns with research into low-resource, zero-shot, or domain-adaptive learning.

\paragraph{Privacy and Data Protection} The integration of contextual user data, even when publicly available, implicates privacy concerns. 
While all data used in this study was publicly accessible at the time of collection, public availability does not equate to ethical permissibility of reuse, especially when data is aggregated or repurposed in ways not anticipated by the original data subjects.

To minimize privacy risks, we applied the following safeguards.
(1) We only publish the tweet IDs and the associated annotations that were collected as part of this study, to be compliant with Twitter’s Developer Agreement and Policy\footnote{\url{https://developer.x.com/en/developer-terms/agreement-and-policy}} and GDPR\footnote{\url{https://eur-lex.europa.eu/eli/reg/2016/679/oj/eng}};
(2) we excluded private, deleted, or protected accounts during data collection;
(3) all experiments were conducted exclusively on secure, institution-hosted servers, ensuring that user information was never transmitted to or processed by any third-party services or external companies.

\paragraph{Informed Consent and Ethical Use} One significant ethical concern is the lack of informed consent: users do not explicitly consent to having their content used for stance modeling, particularly not in ways that infer their beliefs or political attitudes. 
While Twitter’s terms permit academic access to public data, we recognize that consent for platform usage is not equivalent to meaningful, informed consent for automated profiling.

To mitigate this, our analysis is conducted at the aggregate level, and no user-level predictions are released or made publicly identifiable. 
The models are evaluated on annotated tweet-level stance, not on identifying the stance of users across topics. 
No profiling or clustering of users by stance or belief is performed.

\section*{Acknowledgements}
We thank Darya Hryhoryeva, Aishik Mandal and German Ortiz for their valuable feedback and X (previously Twitter) for providing access to the historic Twitter API from 2022 until 2023.
We further acknowledge the authors of the datasets we used for providing them publicly.
We gratefully acknowledge the support of Microsoft with a grant for access to OpenAI GPT models via the Azure cloud (Accelerate Foundation Model Academic Research). 
This work has been funded by the German Federal Ministry of Education and Research (BMBF) under the promotional reference 01UP2229B (KoPoCoV).

\bibliography{anthology,custom}

\appendix

\section{Appendix}
\label{sec:appendix}

\subsection{Data Collection}
\label{app:sec:datacollection}

\subsubsection{covid19-de}

\paragraph{Data Crawling}
% crawling
Besides using 61 different queries for keyword search, we also included tweets from a set of pre-specified German Twitter accounts, such as politicians, journalists, healthcare and research institutes, and news outlets.
These accounts were manually curated and verified, partially based on existing lists from various sources:
\begin{itemize}
    \item \href{https://osf.io/wn48y/}{Twitter- und Facebook-Accounts der Kandidierenden zur Bundestagswahl 2021 by Open Science Framework}
    \item \href{https://www.degruyterbrill.com/document/isbn/9783110769548/html?lang=de}{Deutscher Hochschulverband (2022/Band 1 Universitäten Deutschland) by De Gruyter Saur} 
    \item \href{https://blog.gesis.org/the-german-federal-election-2021-twitter-dataset/}{The German Federal Election 2021 Twitter Dataset by GESIS}
    \item \href{https://osf.io/mqhgp/}{Social media sharing of political elites: An asymmetric American exceptionalism (2022) by Open Science Framework}
    \item \href{https://search.gesis.org/research_data/SDN-10.7802-2415?doi=10.7802/2415}{EPINetz Twitter Politicians (2021) by GESIS Data Archive}
    \item \href{https://search.gesis.org/research_data/ZA6926?doi=10.4232/1.12992}{Monitoring sozialer Medien im Bundestagswahlkampf 2017 by GESIS}
\end{itemize}

Additionally, we leveraged Twitter's internal tweet annotations\footnote{For more information, please refer to the official Twitter documentation at \url{https://developer.x.com/en/docs/x-api/annotations/overview}} to identify tweets related to Covid-19 governmental measures.
Twitter categorizes tweets based on their semantic content.
We collected tweets which have been assigned to the category "COVID-19"\footnote{The corresponding label identifier is 123.1220701888179359745} by Twitter's internal annotation system.
We ignored all retweets.
The initial datasets contains 33,846,159 tweets.
Our preprocessing pipeline includes the elimination of duplicate tweets as well as tweets with less than 5 words.
Further, we removed tweets which were classified as non-German using a pretrained language classifier\footnote{We used a \href{https://huggingface.co/papluca/xlm-roberta-base-language-detection}{XLM-RoBERTa} model for language detection.}.
This resulted in a dataset of 23,972,533 German tweets about the COVID-pandemic. 

In compliance with Twitter’s Developer Agreement and Policy\footnote{\url{https://developer.x.com/en/developer-terms/agreement-and-policy}} and the General Data Protection Regulation (GDPR)\footnote{\url{https://eur-lex.europa.eu/eli/reg/2016/679/oj/eng}}, we only publish the tweet IDs and the associated annotations that were collected as part of this study.
The data is published under a custom license note to be compliant with both Twitter’s Developer Agreement and research-focused use under GDPR.

\paragraph{Annotation}

We recruited seven German-speaking university students as voluntary annotators.
We used the Inception annotation platform~\citep{klie-etal-2018-inception} for the tweet-level annotations.
In the following, we provide an anonymized excerpt of the annotation guidelines, translated from German to English:

\textbf{Objective of the study}
In this collaborative project, we are measuring consensus and polarization in the positions of various social groups (academia, politics, media, the general public) regarding measures to combat the COVID-19 pandemic on the social network Twitter. 
Using innovative methods from the field of Natural Language Processing (NLP), opinions will be automatically captured, and opinion dynamics will be statistically modeled using time-series analysis methods to identify the causes and developments of social polarization processes. 
Specifically, the project will address the following questions, among others: How did various social groups (e.g., politicians or the media) and subgroups (e.g., different political parties and media outlets with varying editorial stances) evaluate the COVID-19 measures, how did this change over time, and how did the positions of the different groups influence one another? 
Since the NLP models developed here can also be applied to future crises, the project allows for the identification of general patterns in the emergence of consensus and polarization during crises and enables a kind of societal early warning system capable of detecting emerging tendencies toward division. 
In addition to this substantive goal, the project also pursues two methodological objectives: First, comparisons of the opinions expressed on Twitter with representative public opinion polls are intended to provide insight into how well the discourse on Twitter serves as an indicator of public opinion. Second, the innovative NLP methods are to be applied to social science research questions and further developed as a result.

\begin{enumerate}
    \item \textbf{Dataset}: The study examines German-language tweets (excluding retweets) posted between January 1, 2020, and March 1, 2022. Access to and storage of the tweets were facilitated via the Twitter Academic API. The tweets were collected using three methods: (1) by keyword (['stayhomesavelifes', 'wirbleibenzuhause', 'infektionsschutz', 'bleibdaheim', 'abstandhalten', \dots]), (2) by topic using automatic classification via Twitter, and (3) by Twitter ID, a manually compiled list of Twitter accounts for all relevant political actors from the fields of politics, media, academia, and civil society, sourced from high-quality sources (Gesis, HBI, BAG Against Hate, publications). The tweets to be coded are provided and coded via the INCEPTION platform.
    \item \textbf{Analysis units}: The units for which a coding sheet is created, are Tweets. Each comment counts as a separate post. Links to photos or other posts are not included in the coding. 
    \item \textbf{Coding units}: Tweets are coded both at the post level and at the level of text segments (e.g., clauses, sentences).
    \item \textbf{Inclusion criteria}: The inclusion criteria determine which posts are coded. On Twitter, all tweets that address government measures to contain the COVID-19 pandemic are coded. The inclusion criteria are explained in more detail below. First, the decision to label a tweet is based on the following question: Does the post actually refer to government measures to contain the COVID-19 pandemic? Please note! The following information should be taken into account when making a decision: (1) A tweet about the pandemic, but no specific measures mentioned. For example, it is possible that a tweet refers to the pandemic but does not mention any specific measures to curb its spread (e.g., \textit{I’ve had COVID-19 four times now \#Sucks}). (2) Measures to contain the pandemic vs. general crisis management measures: Only measures aimed at directly containing the pandemic should be considered, not all crisis management measures in general (e.g., \textit{Where are the grants for schools and educational institutions?}. (3) Government measures vs. measures taken by other institutions/organizations: During the crisis, numerous measures were taken that were initiated by private actors such as companies and organizations. These non-governmental measures are not to be taken into account (e.g., \textit{Our employees will be working from home until the end of the year \#staysafe}). (4) The tweet generally refers to “measures” without providing further details: In this context, we generally assume that these are government measures aimed at containing the pandemic (e.g., \textit{Due to the ongoing COVID-19 measures, the university remains in emergency mode for the time being}). (5) The analysis also takes into account measures implemented by governments in all countries, not just those in Germany (e.g., \textit{\#coronavirus: Italy locks down cities}).
    \item \textbf{Guidelines for coding and data entry}:  (1) As a general rule, if any uncertainties arise during the coding process, please contact one of the project leaders to clarify any questions. The primary goal is not to code as independently as possible, but rather to ensure that the coding is as reliable and valid as possible. (2) For coding, we use the Inception software as our coding platform. The coding can be completed directly on the computer. (3) Each coder is assigned specific tweets whose content is to be coded on specific days (see coding schedule). The relevant tweets are available via MS Teams. (4) For each tweet to be coded, Inception is opened to start the coding process. (5) During coding, only the relevant categories and items are coded. A "zeroing out" (no coding of items in a category) is interpreted as "category does not exist." (6) You must review or view a post as many times as necessary until the coding is complete. (7) If a post has been coded incorrectly: Please reopen Inception and start the coding process over. \item \textbf{General Note}: Please be aware that tweets may contain malicious, suggestive, offensive, or potentially sensitive content. You can pause the annotation at any time and resume it later. 
    \item \textbf{A special feature of annotating Twitter hashtags}: Hashtags are often ambiguous and can only be understood within their specific context. Therefore, the following should be kept in mind when annotating: Hashtags are only considered as context for what is said; they never stand alone. Hashtags are used to determine whether a measure is being addressed. For this, the hashtag must include a measure. Hashtags can be used to provide context and clarify the context of a tweet.

\end{enumerate}

\paragraph{Semi-Automatic Actor Classification}

We compiled comprehensive curated and classified lists of almost 15,000 relevant German Twitter accounts across political, media and science domains to enable systematic collection of elite discourse and accurate actor classification. 
These lists include account handles, user names and respective classification, such as political party, media type, or scientific field of study. 
The lists were assembled from multiple scientific sources including institutional repositories (e.g., GESIS, HBI, Bundespressekonfernez, BAG “Hass im Internet”, Hochschullehrerverzeichnis), academic datasets from studies (e.g., \citet{hohner2022solidarity}, \citet{schwaiger2022gegen}, \citet{stier2020integrating}), and manually curated expert directories (e.g., Hamburger Wahlbeobachter, DPA members) .
Political accounts (n=14,584 before deduplication) encompass German candidates for federal elections, current and former members of the Bundestag, state parliaments, and German members and parties of the European Parliament, federal and state ministries, political parties and their regional affiliates, youth organizations, parliamentary groups, and social movements. 
Media accounts (n=3,557 before deduplication) cover newspapers and magazines, public and private radio and television stations, news programs and editorial offices, journalists accredited to the Bundespressekonferenz, news agencies, and alternative media outlets.
Scientific accounts (n=1,247) included physicians, virologists, epidemiologists, public health experts, and researchers from disciplines relevant to pandemic discourse. 
These accounts were identified through systematic searches of the Hochschullehrer Verzeichnis 2022 (German University Teachers Directory), which catalogs approximately 60,000 academics at German universities including professors, junior professors, and other habilitates with information on their disciplines and institutional affiliations. 
We focused on disciplines related to medicine, virology, epidemiology, public health, and related natural and social sciences.

The classification algorithm proceeded in four sequential steps. 
First, for each account, we preprocessed the biography text by removing special characters and standardizing formatting. 
The cleaned bio was then tokenized into individual words and bi-grams. 
Second, we checked whether the account was included in our curated lists of classified accounts from political, media, or science actors. 
Accounts matching these lists were assigned to the corresponding category. 
Third, for accounts not found in the curated lists, we checked whether any Twitter usernames from our curated lists (prefixed with @) appeared in the account's biography. For example, an account including \textit{@spdde} in its bio would be assigned to the group of political actors. 
Fourth, for accounts without matches in the first two steps, we matched the twitter biographies against predefined keyword lists developed for each actor category. 
For example, keywords for scientists included terms like "researcher," "virologist," and institutional affiliations, while keywords for journalists included "reporter," "editor," and media organization names.
When multiple matches occurred, the algorithm prioritized the category with the strongest match based on the step at which the match was found (curated list > handle mention > keywords) and keyword frequency. 
Accounts with no matches at any step were classified as "citizens" by default, representing the general public rather than institutional actors. 

This semi-automatic approach enabled classification of the entire corpus while maintaining accuracy through the hierarchical matching procedure.

% difference to existing dataset
%Please note that this dataset is different from the one introduced in \citet{beck-etal-2021-investigating}.
%While both datasets are equal in terms of the topic (i.e., Governmental measures to contain the spread of Covid-19) and language (i.e., German), the data instances and annotations are different.
%The focus of the dataset in \citet{beck-etal-2021-investigating} was on making the annotation process more efficient for non-expert annotators.
%In contrast, the goal for creating the dataset in this chapter was to have high-quality annotations by annotators who received substantial annotation training and quality control.
%This enables us to investigate the impact of different context features on the model performance, while the annotation quality is kept constant.
%Further, we used a extensively larger crawl of Twitter posts over a larger time period (> 2 years) to create the dataset in this chapter.
%This enables us to cover a broader range of aspects (e.g., vaccination) and opinion dynamics regarding governmental measures during the Covid-19 crisis.

\subsection{Data Cleaning}
\label{app:sec:datacleaning}

Due to the dynamic nature of Twitter and its users, posts get deleted, removed, or set to private. 
As we collect the metadata post-hoc of initial dataset retrieval, a large percentage cannot be retrieved anymore. 
To increase the transparency of our data collection procedure, 
we analyzed the changes in class (im)balance due to filtering and present the results in the \Cref{app:tab:datasets_filtered}. 
We observe that label distribution remains comparable, except for the labels \textit{favor} and \textit{against} for dataset \texttt{covid19glandt} which differ by 10pp each, compared to the original dataset size.

\begin{table*}[t]
\centering
\small
\setlength{\tabcolsep}{4pt}
\begin{tabular}{llrl}
\toprule
\textbf{Dataset} & & \textbf{Size} & \textbf{Label Distribution} \\
\midrule
\multirow{2}{*}{covid19beck}   & initial  & 3,750  & neutral (80\%), positiv (10\%), negativ (10\%) \\
                               & filtered & 3,046  & neutral (80\%), positiv (10\%), negativ (9\%) \\
\midrule
\multirow{2}{*}{covid19glandt} & initial  & 6,133  & favor (34\%), none (34\%), against (32\%) \\
                               & filtered & 1,017  & favor (42\%), none (34\%), against (22\%) \\
\midrule
\multirow{2}{*}{semeval2016}   & initial  & 4,163  & against (49\%), favor (25\%), none (26\%) \\
                               & filtered &   859  & against (44\%), favor (28\%), none (26\%) \\
\midrule
\multirow{2}{*}{wtwt}          & initial  & 51,284 & comment (41\%), unrelated (38\%), support (13\%), refute (8\%) \\
                               & filtered & 9,263  & comment (39\%), unrelated (38\%), support (12\%), refute (9\%) \\
\bottomrule
\end{tabular}
\caption{Dataset sizes and label distributions before and after filtering.}
\label{app:tab:datasets_filtered}
\end{table*}

\paragraph{Covid19-Glandt}

\begin{table}[h!]
    \centering
    \resizebox{.49\textwidth}{!}{%
    \begin{tabular}{|l|l|}
    \hline
    \textbf{Original Value} & \textbf{Replaced Value} \\ \hline
    face\_masks & Face masks \\ \hline
    stay\_at\_home\_orders & Stay at home orders \\ \hline
    school\_closures & School closures \\ \hline
    fauci & Dr. Anthony Fauci \\ \hline
    \end{tabular}
    }
    \caption{Conversion of original values to human-readable values for topics in the Covid19-Glandt dataset.}
    \label{app:tab:covid19glandt-topics}
\end{table}

% excluding noisy data
The original Covid19-Glandt dataset contains datafiles with noisy data (i.e., the filenames contain \textit{noisy}).
We did not consider these files for our experiments.

% mapping the source topics to human-readable topics
The original dataset contains topics represented by abbreviations, making them not immediately interpretable. 
To address this, we manually mapped these topics to human-readable labels, drawing from their descriptions in the source publication. 
The complete mapping is provided in \cref{app:tab:covid19glandt-topics}.

\paragraph{WTWT}

\begin{table}[h!]
    \centering
    \resizebox{.49\textwidth}{!}{%
    \begin{tabular}{|l|l|}
    \hline
    \textbf{Original Value} & \textbf{Replaced Value} \\ \hline
    AET\_HUM & Aetna acquires Humana \\ \hline
    ANTM\_CI & Anthem acquires Cigna \\ \hline
    CVS\_AET & CVS Health acquires Aetna \\ \hline
    CI\_ESRX & Cigna acquires Express Scripts \\ \hline
    FOXA\_DIS & Disney acquires 21st Century Fox \\ \hline
    \end{tabular}
    }
    \caption{Conversion of original values to human-readable values for topics in the WTWT dataset.}
    \label{app:tab:wtwt-topics}
\end{table}

% mapping the source topics to human-readable topics
Similar to the Covid19-Glandt dataset, the original WTWT~\citep{conforti-etal-2020-will} dataset contains topics that are abbreviated and thus not directly interpretable.
We provide the mapping in \cref{app:tab:wtwt-topics}.

\subsection{Twitter Metadata and LLM-Generated Context}
\label{app:sec:twittermetadata}

We provide the prompt templates used to generate context (i.e., \textit{Target Description} and \textit{Tweet-Specific Context}) in \Cref{ch:prompting_noisy_domains:tab:prompts}. 

\Cref{ch:prompting_noisy_domains:tab:twitter_metadata:first} and \Cref{ch:prompting_noisy_domains:tab:twitter_metadata:second} provide examples of the retrieved metadata and LLM-generated context for each dataset\footnote{Note that we replaced or slightly changed non-public usernames for privacy reasons. Further, we replaced urls with a placeholder}.

\begin{table*}[t]
      \small
  
      \centering
      \begin{tabular}{p{2.2cm}p{5.1cm}p{5.1cm}}
      \toprule
      \textbf{Dataset} & \textbf{covid19-de} & \textbf{covid19-glandt}  \\                                                                                   
      \midrule
      \textbf{Target} & Regierungs-maßnahmen zu Covid-19 & School closures \\
  
      \textbf{Tweet} & Die COVID-19-Impfung schützt gut vor Hospitalisierung und Tod. Jetzt Impftermin vereinbaren! [URL] \#bayerngemeinsam \#coronavirus \#impfung [URL]  &
  @GovMurphy Your lock downs are all about politics. We all sacrificed to flatten the curve. Open up NJ now. If masks work or being 6 feet apart works then NJ can open schools
  - right. Or here is another thought. \\
  
      \textbf{Stance} & pro & against \\
  
      \textbf{Username} & Staatsministerium für Gesundheit und Pflege Bayern & HeadCoachHurst \\
  
      \textbf{User Description} & Bayerisches Staatsministerium für Gesundheit und Pflege - Offizieller Twitter Account. Informationen zu Gesundheit und Pflege in Bayern und
  zur Corona-Pandemie. & Husband, Dad, Head Football Coach and Athletic Director at Pepperell High School \\
  
      \textbf{Other Tweet by User} & -- & RT \@BarrettSallee: Schools opening - in a responsible way - should not be a political issue at all. Kids, especially elementary
  schoolers, [\dots]  \\
  
  schoolers, [\dots]  \\

      \textbf{Target Description} & Die Regierungsmaßnahmen zu Covid-19 umfassten eine Vielzahl von Strategien, die darauf abzielten, die Ausbreitung des Virus zu kontrollieren
   und die öffentliche Gesundheit zu schützen. Zu den wichtigsten Maßnahmen gehörten Ausgangsbeschränkungen, Kontaktverbote, [\dots] & School closures refer to the temporary or
   permanent shutdown of educational institutions, often in response to emergencies, public health crises, or other significant circumstances. Notably highlighted during the
  COVID-19 pandemic, such closures can impact students' learning, emotional well-being, and social development. They have broader implications for families, communities, and
  the economy, [\dots] \\
  
      \textbf{Tweet-Specific Context} & Der Tweet bezieht sich auf die COVID-19-Impfung und hebt deren Schutzwirkung vor schweren Verläufen wie Hospitalisierung und Tod hervor.
   In der Zeit der COVID-19-Pandemie haben Regierungen weltweit, einschließlich in Deutschland, Impfkampagnen gestartet, um die Bevölkerung [\dots] & The tweet addresses school
   closures in New Jersey during the COVID-19 pandemic, reflecting concerns about the extended lockdown measures and their perceived political motivations. The concept of
  ``flattening the curve'' refers to efforts made in early 2020 to reduce the COVID-19 infection rate to prevent overwhelming healthcare systems. The commenter is questioning
  the continued closures of schools, [\dots] \\

  \textbf{Author Class} & political actors & -- \\

      \textbf{Author Group} & State political actors & -- \\

      \bottomrule
      \end{tabular}
  
      \caption{Examples of Twitter Metadata for each of the datasets used in our experiments. We retrieve the \textit{Username}, \textit{User Description}, and \textit{Other
  Tweets by User} for each tweet using the official Twitter API. We further generate additional contextual information by prompting GPT-4o for a \textit{Target Description} and
   a \textit{Tweet-Specific Context}. The \textit{Author Class} and \textit{Author Group} are dataset-specific annotations available only for \texttt{covid19-de}. We abbreviated some of the features due to space limitations.}
      \label{ch:prompting_noisy_domains:tab:twitter_metadata:first}
  \end{table*}

\begin{table*}[t]
    \small
    \centering
    \begin{tabular}{p{2.2cm}p{5.1cm}p{5.1cm}}
    \toprule
    \textbf{Dataset} & \textbf{semeval2016t6} & \textbf{wtwt} \\
    \midrule
    \textbf{Target} & Legalization of Abortion & CVS Health acquires Aetna \\

    \textbf{Tweet} & What is it that makes the human race behave in an undignified way to gain attention? \#furgison \#prideparade \#potus \#isis \#lgbt & CVS and Aetna are reportedly nearing a \$66 billion deal (CVS, AET) \#healthcare [URL] - Grow you \\

    \textbf{Stance} & none & support \\

    \textbf{Username} & DrMIDI &  DrugChannelInfo \\

    \textbf{User Description} & American Composer/Music Producer/Recording Artist/Entertainer/Professor of Music Technology & Expert insights from [NAME] on pharmaceutical economics and the drug distribution system. Contact me at [EMAIL] \\

    \textbf{Other Tweet by User} &  I liked a \@YouTube video [URL] The Narrow Way Or The High Way: Are You Striving To Enter or Resting in Salvation? &  Rumor time: \@CVSHealth \$CVS / \@Aetna \$AET deal will dodge vertical \#antitrust challenge from \@TheJusticeDept - major issue may be horizontal \#Medicare \#PartD issues [URL] \\

    \textbf{Target Description} & The legalization of abortion refers to the process of making abortion lawful, allowing women the right to terminate their pregnancies under certain conditions [\dots] & CVS Health's acquisition of Aetna, completed in November 2018, was a landmark merger in the healthcare industry. The deal, valued at approximately \$69 billion, aimed to combine CVS's extensive pharmacy and retail operations with Aetna's health insurance services [\dots] \\

    \textbf{Tweet-Specific Context} & The tweet reflects a general frustration with human behavior, particularly when seeking attention through controversial or provocative means. In this context of the legalization of abortion, the hashtags—such as \#prideparade [\dots] & The tweet references a discussion on the implications of major mergers among healthcare payers, citing expert opinions. Aetna's planned acquisition of Humana, which was part of a larger trend of consolidation [\dots] \\

    \bottomrule
    \end{tabular}
    \caption{Examples of Twitter Metadata for each of the datasets used in our experiments. We retrieve the \textit{Username}, \textit{User Description}, and \textit{Other Tweets by User} for each tweet using the official Twitter API. We further generate additional contextual information by prompting GPT-4o for a \textit{Target Description} and a \textit{Tweet-Specific Context}. We abbreviated some of the features due to space limitations.}
    \label{ch:prompting_noisy_domains:tab:twitter_metadata:second}
\end{table*}

\begin{table*}[t]
	\centering
	\begin{tabular}{p{3cm}p{3cm}p{8cm}}
		\toprule
		\textbf{Context} & \textbf{Datasets} & \textbf{Prompt Template} \\
		\midrule
		\multirow{2}{3cm}{Target Description} & covid19-glandt, semeval2016t6, wtwt & Describe the topic '\{target\}' shortly. \\
														& covid19-de                        & Fasse das Thema '\{target\}' kurz in einem Text zusammen. \\
		\midrule
		\multirow{2}{3cm}{Tweet-Specific Context} & covid19-glandt, semeval2016t6, wtwt & Provide context for the following tweet regarding the given topic where possible. Do not interpret the tweet, just provide contextual information to better understand it. Keep it short and concise. Tweet: '\{text\}' Topic: '\{target\}' \\
														& covid19-de                        & Beschreibe den Kontext für den folgenden Tweet in Bezug auf das angegebene Thema. Interpretiere den Tweet nicht, sondern stelle nur kontextuelle Informationen bereit, um das Verständnis zu erleichtern. Halte es kurz und prägnant. Tweet: '\{text\}' Thema: '\{target\}' \\														   
		\bottomrule
	\end{tabular}
	\caption{Prompt templates for generating additional context information for stance detection in Twitter.}
	\label{ch:prompting_noisy_domains:tab:prompts}
\end{table*}

\subsection{Prompt Configuration}
\label{app:sec:models}

\subsubsection{Prompt Instructions}
  \label{appendix:prompt-instructions}

All prompts are built from a shared instruction template, instantiated in English for \texttt{semeval2016}, \texttt{wtwt}, and \texttt{covid19glandt}, and in German for the \texttt{covid19beck} dataset. 
Each prompt is finalized with the tweet text and a constrained stance slot, \texttt{Stance (\{stance\_labels\}):}, whose value is decoded under a JSON schema restricting the answer to the dataset's label set.

\paragraph{Zero-Shot and Few-Shot.}
The zero-shot template instructs the model to act as a stance-labelling assistant and to predict the stance toward a given target. 
In its no-context form, the English instruction reads:
\begin{quote}\itshape
You are a stance-labelling assistant.\\
Given the tweet text, what stance is expressed towards `\{target\}'?
\end{quote}
The German variant is:
\begin{quote}\itshape
Du sollst helfen die Haltung in Tweets zu identifizieren.\\
Gegeben dem Tweet, welche Haltung wird gegenüber dem Thema `\{target\}'
ausgedrückt?
\end{quote}
When one or more contextual features are added, ``Given the tweet text''
(resp. ``Gegeben dem Tweet'') is replaced by ``Given the context of a tweet
and the tweet text'' (resp. ``Gegeben den Kontextinformationen des Tweets und
dessen Text''), and the corresponding context blocks (username, user
biography, other tweets by the user, target description, tweet-target
context) are prepended.

For the few-shot setting, $n$ examples are drawn from the same dataset using stratified sampling over labels (with the test instance excluded to prevent leakage) and rendered with the same context blocks as the query. 
The example block is prefaced by ``Here are examples of Twitter texts with contextual information and their stance towards a given target:'' (resp. the analogous German phrasing) and prepended to the zero-shot instruction.

\paragraph{Chain-of-Thought (CoT).}
The CoT setting reuses the default zero-shot template and elicits explicit reasoning by appending a step-by-step trigger directly after the stance slot, i.e. \texttt{Stance (\{stance\_labels\}): let's think step by step.} in English and the analogous \emph{``Lass uns Schritt für Schritt denken.''} in German. Decoding remains constrained by the same JSON schema, so the final emitted token is a label from the dataset's stance inventory.

\paragraph{LLM-as-a-Judge.}
The LLM-as-Judge strategy is a two-stage pipeline: an initial zero-shot prediction is post-hoc verified by a second LLM call. 
The judge receives the tweet text, the target, and the candidate stance, and is instructed to decide whether the label is \emph{coherent} with the text's attitude toward the target, returning either the original label or the corrected one. 
Output is constrained via a Pydantic schema with fields \texttt{answer}, \texttt{reasoning}, and \texttt{implied\_stance}, parsed with the \texttt{instructor} library. 
The English judge instruction follows the schema below:
\begin{quote}\itshape
You are a stance-labelling assistant.\\
\textbf{Task:} Determine whether the provided stance label correctly
represents the text's attitude or opinion toward the target. A stance is
considered coherent if it accurately reflects the text's implied or
explicit position toward the target.\\
\textbf{Guidelines:} If the text expresses a clear opinion in favor of the
target, the coherent stance is `favor'; if against, `against'; if no
clear opinion is expressed, `none'.\\
\textbf{Steps:} (1) Think step by step and analyse the text and its
context to determine its true stance toward the target. (2) Compare it
with the given stance label. (3) Provide the same stance if the given
label is coherent, otherwise provide the actual stance.
\end{quote}
The German variant uses the corresponding label inventory
(\textit{positiv}, \textit{negativ}, \textit{neutral}) and the analogous
guidelines. 
As with the other strategies, the final answer is decoded
under a JSON schema restricted to the dataset's stance labels.

\subsubsection{Structured Output}

The models are hosted using the vLLM\footnote{\url{https://vllm.ai/}} inference service on a cluster with mixed GPUs (i.e., NVIDIA H100 SXM5 and NVIDIA RTX Pro 6000 using CUDA 13.0).
We provide an example prompting configuration (Listing~\ref{lst:prompt_conf}) for the structured outputs framework using JSON which is used to force reasoning models to generate predictions which adhere to each dataset's stance classes.

\begin{lstlisting}[
  language=Python,
  caption=Prompting example with the structured output configuration of vLLM,
  basicstyle=\scriptsize\ttfamily,
  breaklines=true,
  breakatwhitespace=false,
  postbreak=\mbox{\textcolor{gray}{$\hookrightarrow$}\space},
  columns=fullflexible,
  showstringspaces=false,
  label={lst:prompt_conf}
]
import json
import requests 
# vLLM's OpenAI-compatible endpoint
base_url = "http://localhost:8000/v1/chat/completions"
model_name = "openai/gpt-oss-20b"
labels = ["FAVOR", "AGAINST", "NONE"]  # allowed answers                                                  messages = [{"role": "user", "content": "Tweet: 'Masks work.'  Target: 'face masks'.  Stance?"}]
payload = {     
  "model": model_name,
  "messages": messages,
  "temperature": 0.5,
  "max_tokens": 4096,
  "seed": 42,
  # Forces the generated text to be exactly one of `labels`.
  # vLLM's guided decoding: {"choice": [...]}  ->  constrained output
  "structured_outputs": {"choice": labels},
  # Optional: keep reasoning traces enabled for thinking-capable models
  "chat_template_kwargs": {"thinking": True},
}
response = requests.post(base_url, json=payload, timeout=150).json()
answer = response["choices"][0]["message"]["content"].strip()
\end{lstlisting}

\subsection{Model Results}
\label{app:sec:results}

We provide the corresponding result tables for models \texttt{Qwen2.5-32B} (Table~\ref{tab:qwen2532b_results_extended}), \texttt{Gemma3-27B} (Table~\ref{tab:gemma327b_results_extended}) and \texttt{Ministral3-14B} (Table~\ref{tab:ministral14b_results_extended}).

\begin{table*}[t!]
    \small
    \centering
    \resizebox{.99\textwidth}{!}{%
    \begin{tabular}{c p{3cm}p{2.5cm}p{2.5cm}p{2.5cm}p{2.5cm}}
    \toprule
 & Setting & covid19-de & covid19-glandt & semeval2016t6 & wtwt \\
    \midrule
    \multirow{4}{*}{\rotatebox{90}{No context}} & Zero-Shot & $48.39_{\pm 1.31}$ & $61.86_{\pm 0.10}$ & $65.56_{\pm 0.20}$ & $45.67_{\pm 0.08}$ \\
     & Few-Shot (n=10) & $47.86_{\pm 2.83}$ & $71.86_{\pm 3.36}$ & $74.00_{\pm 0.61}$ & $\mathbf{63.95}_{\pm 2.03}$ \\
     & CoT & $23.02_{\pm 3.64}$ & $13.35_{\pm 0.90}$ & $21.16_{\pm 0.84}$ & $9.59_{\pm 6.63}$ \\
     & LLM-as-a-Judge & $\mathbf{51.98}_{\pm 0.37}$ & $54.32_{\pm 0.15}$ & $62.32_{\pm 0.16}$ & $51.35_{\pm 0.33}$ \\
    \midrule
    \multirow{6}{*}{\rotatebox{90}{\shortstack{Zero-Shot \\ + Context}}} & Username & $38.78_{\pm 2.94}$ & $64.96_{\pm 0.20}$$^{*}$ & $65.30_{\pm 0.64}$ & $42.41_{\pm 0.20}$ \\
     & User Description & $42.55_{\pm 2.58}$ & $63.74_{\pm 0.26}$$^{*}$ & $65.79_{\pm 0.39}$ & $41.05_{\pm 0.19}$ \\
     & Other Tweets & - & $65.28_{\pm 0.20}$$^{*}$ & $62.11_{\pm 0.43}$ & $40.62_{\pm 0.19}$ \\
     & Target Description & $45.87_{\pm 2.64}$ & $63.62_{\pm 0.21}$$^{*}$ & $71.71_{\pm 0.32}$$^{*}$ & $50.25_{\pm 0.15}$$^{*}$ \\
     & Tweet-Specific Context & $43.55_{\pm 2.28}$ & $52.74_{\pm 0.12}$ & $60.07_{\pm 0.14}$ & $36.74_{\pm 0.30}$ \\
     & All & - & $58.96_{\pm 0.75}$ & $59.90_{\pm 0.17}$ & $38.61_{\pm 0.24}$ \\
    \midrule
    \multirow{6}{*}{\rotatebox{90}{\shortstack{Few-Shot \\ + Context}}} & Username & $42.21_{\pm 3.69}$ & $\mathbf{72.75}_{\pm 2.55}$ & $73.97_{\pm 0.12}$ & $63.47_{\pm 1.29}$ \\
     & User Description & $45.91_{\pm 5.96}$ & $72.39_{\pm 2.21}$ & $72.92_{\pm 0.68}$ & $59.28_{\pm 4.06}$ \\
     & Other Tweets & - & $72.50_{\pm 1.79}$$^{*\dag}$ & $68.20_{\pm 0.70}$ & $57.35_{\pm 2.53}$ \\
     & Target Description & $50.62_{\pm 3.25}$$^{*}$ & $71.77_{\pm 1.90}$ & $\mathbf{76.13}_{\pm 2.34}$$^{*\dag}$ & $59.25_{\pm 0.89}$ \\
     & Tweet-Specific Context & $44.41_{\pm 3.20}$ & $62.10_{\pm 1.06}$ & $65.19_{\pm 1.80}$ & $51.96_{\pm 0.23}$ \\
     & All & - & $65.42_{\pm 1.85}$ & $64.96_{\pm 2.02}$ & $49.34_{\pm 0.94}$ \\
    \bottomrule
    \end{tabular}
    }
    \caption{Macro F1 results for \texttt{Qwen2.5-32B}. Bold results are best per dataset, $^{*}$ is for stat. significant better performance than the same-reasoning-type baseline (zero-shot baseline for the upper metadata block, few-shot baseline for the lower metadata block), and $^{\dag}$ is for stat. significant better performance than the other reasoning-type baselines (zero-shot block: vs.\ Few-Shot, CoT, Judge; few-shot block: vs.\ CoT, Judge) using a paired bootstrap test ($p < 0.05$).}
    \label{tab:qwen2532b_results_extended}
\end{table*}

\begin{table*}[t!]
    \small
    \centering
    \resizebox{.99\textwidth}{!}{%
    \begin{tabular}{c p{3cm}p{2.5cm}p{2.5cm}p{2.5cm}p{2.5cm}}
    \toprule
 & Setting & covid19-de & covid19-glandt & semeval2016t6 & wtwt \\
    \midrule
    \multirow{4}{*}{\rotatebox{90}{No context}} & Zero-Shot & $38.25_{\pm 0.73}$ & $54.92_{\pm 0.21}$ & $72.47_{\pm 0.27}$ & $45.32_{\pm 0.11}$ \\
     & Few-Shot (n=10) & $43.00_{\pm 1.18}$ & $66.51_{\pm 2.32}$ & $70.52_{\pm 1.65}$ & $\mathbf{64.09}_{\pm 1.47}$ \\
     & CoT & $42.06_{\pm 0.54}$ & $28.04_{\pm 2.14}$ & $23.53_{\pm 1.71}$ & $27.52_{\pm 2.37}$ \\
     & LLM-as-a-Judge & $46.70_{\pm 0.39}$ & $41.77_{\pm 0.73}$ & $25.75_{\pm 0.61}$ & $44.69_{\pm 1.19}$ \\
    \midrule
    \multirow{6}{*}{\rotatebox{90}{\shortstack{Zero-Shot \\ + Context}}} & Username & $37.11_{\pm 0.63}$ & $54.75_{\pm 0.36}$ & $71.16_{\pm 0.09}$ & $46.45_{\pm 0.09}$$^{*}$ \\
     & User Description & $37.85_{\pm 0.67}$ & $55.45_{\pm 0.35}$ & $70.63_{\pm 0.11}$ & $45.82_{\pm 0.16}$ \\
     & Other Tweets & - & $58.13_{\pm 0.05}$$^{*}$ & $62.91_{\pm 0.13}$ & $47.23_{\pm 0.04}$$^{*}$ \\
     & Target Description & $39.11_{\pm 0.90}$ & $57.62_{\pm 0.35}$$^{*}$ & $\mathbf{75.69}_{\pm 0.09}$$^{*\dag}$ & $51.31_{\pm 0.03}$$^{*}$ \\
     & Tweet-Specific Context & $34.03_{\pm 0.56}$ & $55.67_{\pm 0.17}$ & $59.66_{\pm 0.09}$ & $44.28_{\pm 0.42}$ \\
     & All & - & $58.01_{\pm 0.48}$$^{*}$ & $59.81_{\pm 0.16}$ & $46.49_{\pm 0.58}$$^{*}$ \\
    \midrule
    \multirow{6}{*}{\rotatebox{90}{\shortstack{Few-Shot \\ + Context}}} & Username & $43.14_{\pm 0.95}$ & $70.06_{\pm 4.91}$$^{*\dag}$ & $72.73_{\pm 0.81}$$^{*\dag}$ & $62.11_{\pm 1.42}$ \\
     & User Description & $43.53_{\pm 1.56}$ & $68.96_{\pm 4.50}$ & $70.42_{\pm 0.72}$ & $56.02_{\pm 2.26}$ \\
     & Other Tweets & - & $66.72_{\pm 0.74}$ & $62.15_{\pm 3.26}$ & $52.67_{\pm 1.65}$ \\
     & Target Description & $\mathbf{46.83}_{\pm 1.64}$$^{*}$ & $\mathbf{71.05}_{\pm 0.87}$$^{*\dag}$ & $75.41_{\pm 1.35}$$^{*\dag}$ & $62.56_{\pm 2.56}$ \\
     & Tweet-Specific Context & $37.57_{\pm 1.48}$ & $68.46_{\pm 1.72}$$^{*\dag}$ & $68.50_{\pm 0.23}$ & $56.14_{\pm 2.93}$ \\
     & All & - & $65.94_{\pm 1.24}$ & $62.83_{\pm 1.87}$ & $49.19_{\pm 2.35}$ \\
    \bottomrule
    \end{tabular}
    }
    \caption{Macro F1 results for \texttt{Gemma3-27B}. Bold results are best per dataset, $^{*}$ is for stat. significant better performance than the same-reasoning-type baseline (zero-shot baseline for the upper metadata block, few-shot baseline for the lower metadata block), and $^{\dag}$ is for stat. significant better performance than the other reasoning-type baselines (zero-shot block: vs.\ Few-Shot, CoT, Judge; few-shot block: vs.\ CoT, Judge) using a paired bootstrap test ($p < 0.05$).}
    \label{tab:gemma327b_results_extended}
\end{table*}

\begin{table*}[t!]
    \small
    \centering
    \resizebox{.99\textwidth}{!}{%
    \begin{tabular}{c p{3cm}p{2.5cm}p{2.5cm}p{2.5cm}p{2.5cm}}
    \toprule
 & Setting & covid19-de & covid19-glandt & semeval2016t6 & wtwt \\
    \midrule
    \multirow{4}{*}{\rotatebox{90}{No context}} & Zero-Shot & $39.99_{\pm 1.00}$ & $55.75_{\pm 2.41}$ & $50.13_{\pm 3.89}$ & $28.98_{\pm 15.74}$ \\
     & Few-Shot (n=10) & $\mathbf{41.06}_{\pm 1.25}$ & $60.88_{\pm 2.76}$ & $65.41_{\pm 2.84}$ & $24.96_{\pm 16.10}$ \\
     & CoT & $29.73_{\pm 0.00}$ & $25.45_{\pm 3.46}$ & $34.16_{\pm 6.20}$ & $8.76_{\pm 5.29}$ \\
     & LLM-as-a-Judge & $34.74_{\pm 1.50}$ & $47.34_{\pm 10.89}$ & $40.40_{\pm 8.41}$ & $24.39_{\pm 9.32}$ \\
    \midrule
    \multirow{6}{*}{\rotatebox{90}{\shortstack{Zero-Shot \\ + Context}}} & Username & $40.11_{\pm 1.20}$ & $58.12_{\pm 2.76}$$^{*}$ & $47.73_{\pm 4.42}$ & $22.38_{\pm 10.56}$ \\
     & User Description & $40.02_{\pm 0.87}$ & $59.73_{\pm 2.06}$$^{*}$ & $53.28_{\pm 3.45}$$^{*}$ & $20.83_{\pm 9.50}$ \\
     & Other Tweets & - & $64.53_{\pm 1.46}$$^{*\dag}$ & $60.07_{\pm 1.94}$$^{*}$ & $23.45_{\pm 8.75}$ \\
     & Target Description & $37.82_{\pm 1.26}$ & $59.64_{\pm 2.20}$$^{*}$ & $56.38_{\pm 2.07}$$^{*}$ & $33.06_{\pm 13.25}$$^{*\dag}$ \\
     & Tweet-Specific Context & $36.57_{\pm 0.77}$ & $57.01_{\pm 0.86}$ & $58.31_{\pm 0.61}$$^{*}$ & $30.78_{\pm 11.34}$ \\
     & All & - & $68.23_{\pm 1.20}$$^{*\dag}$ & $56.88_{\pm 0.42}$$^{*}$ & $33.14_{\pm 8.74}$ \\
    \midrule
    \multirow{6}{*}{\rotatebox{90}{\shortstack{Few-Shot \\ + Context}}} & Username & $40.63_{\pm 1.09}$ & $59.98_{\pm 3.31}$ & $66.53_{\pm 1.11}$$^{*\dag}$ & $31.63_{\pm 13.28}$$^{*\dag}$ \\
     & User Description & $39.56_{\pm 4.42}$ & $60.06_{\pm 4.96}$ & $65.02_{\pm 1.14}$ & $17.61_{\pm 9.64}$ \\
     & Other Tweets & - & $69.36_{\pm 3.30}$$^{*\dag}$ & $61.19_{\pm 6.38}$ & $24.24_{\pm 13.78}$ \\
     & Target Description & $39.98_{\pm 2.20}$ & $61.47_{\pm 3.13}$ & $\mathbf{68.94}_{\pm 2.78}$$^{*\dag}$ & $33.80_{\pm 7.59}$$^{*\dag}$ \\
     & Tweet-Specific Context & $39.01_{\pm 2.50}$ & $59.48_{\pm 2.36}$ & $63.54_{\pm 1.81}$ & $25.80_{\pm 10.97}$ \\
     & All & - & $\mathbf{72.89}_{\pm 1.56}$$^{*\dag}$ & $65.55_{\pm 0.63}$ & $\mathbf{47.90}_{\pm 7.19}$$^{*\dag}$ \\
    \bottomrule
    \end{tabular}
    }
    \caption{Macro F1 results for \texttt{Ministral3-14B}. Bold results are best per dataset, $^{*}$ is for stat. significant better performance than the same-reasoning-type baseline (zero-shot baseline for the upper metadata block, few-shot baseline for the lower metadata block), and $^{\dag}$ is for stat. significant better performance than the other reasoning-type baselines (zero-shot block: vs.\ Few-Shot, CoT, Judge; few-shot block: vs.\ CoT, Judge) using a paired bootstrap test ($p < 0.05$).}
    \label{tab:ministral14b_results_extended}
\end{table*}

\subsubsection{covid19-de context feature}

\Cref{ch:finalch:tab:semisupervised_overall_delta} reports the overall macro-F1 change relative to the no-context baseline for all four evaluated LLMs and both author features. The picture is consistent with the per-class breakdown for \texttt{GPT-OSS-20B} in \Cref{ch:finalch:tab:semisupervised_perclass_gptoss20b}; 15 of 16 (model, setting, feature) cells regress, whereas the \texttt{GPT-OSS-20B}/FS/AC at $+0.32$\,pp, is well within seed variance.
The magnitude of regression varies by model.
\texttt{GPT-OSS-20B} shows the smallest drops (mostly below $1$\,pp), \texttt{Ministral3-14B} and \texttt{Gemma3-27B} cluster between $1$ and $2$\,pp, and \texttt{Qwen2.5-32B} exhibits the largest regressions, peaking at $-5.95$\,pp under zero-shot \textit{Author Class}. 
Within each model, \textit{Author Class} (AC) and \textit{Author Group} (AG) yield broadly similar $\Delta$F1, with no consistent advantage for the finer-grained group identity over the coarse 4-way class. 
Taken together, the regression observed for \texttt{GPT-OSS-20B} in the main text is not a model-specific artifact. 
Across all four LLMs, neither author-level annotation translates into measurable F1 gains for stance detection. 

\begin{table}[t!]
    \small
    \centering
    \begin{tabular}{llcccc}
    \toprule
    Setting & Feat. & \texttt{Ministral3-14B} & \texttt{Gemma3-27B} & \texttt{Qwen2.5-32B} & \texttt{GPT-oss-20B} \\
    \midrule
    \multirow{2}{*}{ZS} & AC & {\color{red}$\downarrow$} -1.42 & {\color{red}$\downarrow$} -2.13 & {\color{red}$\downarrow$} -5.95 & {\color{red}$\downarrow$} -0.90 \\
     & AG & {\color{red}$\downarrow$} -1.66 & {\color{red}$\downarrow$} -1.73 & {\color{red}$\downarrow$} -4.46 & {\color{red}$\downarrow$} -1.21 \\
    \midrule
    \multirow{2}{*}{FS} & AC & {\color{red}$\downarrow$} -1.11 & {\color{red}$\downarrow$} -1.66 & {\color{red}$\downarrow$} -1.81 & {\color{green}$\uparrow$} +0.32 \\
     & AG & {\color{red}$\downarrow$} -0.17 & {\color{red}$\downarrow$} -1.83 & {\color{red}$\downarrow$} -2.05 & {\color{red}$\downarrow$} -0.50 \\
    \bottomrule
    \end{tabular}
    \caption{Overall macro-F1 change ($\Delta$, in pp) over the no-context baseline when injecting author annotations on \texttt{covid19-de}. Settings: zero-shot (ZS), few-shot (FS); features: author class (AC), author group (AG). 15 of 16 cells regress; the only non-negative cell is \texttt{GPT-oss-20B}/FS/AC at $+0.32$\,pp.}
    \label{ch:finalch:tab:semisupervised_overall_delta}
\end{table}
\subsection{Analysis}

\subsubsection{LLM-as-a-Judge analysis}
\label{app:judge_analysis}

To understand why \textit{LLM-as-a-Judge} is particular impactful on \texttt{covid19-de} but not elsewhere, we decompose the Judge's output against its underlying zero-shot (ZS) prediction into five disjoint cells. % (\Cref{ch:finalch:tab:judge_flip_decomposition}).
The Judge either \emph{agrees} with the ZS label (correctly or wrongly), \emph{corrects} a wrong ZS prediction to the gold label, \emph{regresses} a correct prediction to a wrong one, or makes a \emph{lateral} wrong-to-wrong change. 
On \texttt{covid19-de}, corrections (25.5\%) outweigh regressions (8.0\%) significantly with three of four models exhibiting net gains between $+18$ and $+28$ percentage points. 
On the remaining three datasets, corrections and regressions are approximately balanced (\texttt{covid19-glandt} $17.2$ vs.\ $19.7$, \texttt{wtwt} $8.3$ vs.\ $9.5$) or actively inverted (\texttt{semeval2016t6} $11.3$ vs.\ $22.6$).
In other words, the Judge intervenes at roughly the same rate but cancels itself out, and on \texttt{semeval2016t6} it actively degrades the baseline. 
The latter effect is amplified by a model-level outlier.
\texttt{Gemma3-27B} regresses on 45\% of \texttt{semeval2016t6} test instances and by elevated lateral-wrong-to-wrong rates from \texttt{Ministral3-14B} on the same dataset (20.1\%) and on \texttt{wtwt} (22.6\%).
This indicates that some judges relabel essentially arbitrarily once the original prediction is incorrect.
  
The per-gold-class breakdown in Table~\ref{ch:finalch:tab:judge_per_class_delta} provides a clearer picture. 
On \texttt{covid19-de}, the entire net gain is concentrated on the \emph{neutral} class, which carries 80.5\% of the test mass and improves by $+15.9$pp, while both stance-bearing classes are negatively impacted. % ($-4.4$ on \emph{positiv}, $-1.7$ on \emph{negativ}). 
The Judge is therefore not a general-purpose calibrator but a directional ``stance-to-neutral'' corrector, and its apparent strength on \texttt{covid19-de} is largely a consequence of the heavy neutral skew in the gold distribution. 
The same mechanism predicts the Judge's failure modes elsewhere. 
On the more class-balanced \texttt{covid19-glandt} and \texttt{semeval2016t6}, all non-trivial stance classes lose F1 % (\emph{against} $-15.0$ and $-21.3$pp, respectively)
, and on \texttt{wtwt} the only significant drop is on the \emph{comment} label %($-8.4$pp)
, offset by small gains on the more committal classes. 
We therefore interpret the \texttt{covid19-de} result not as evidence that Judge-style self-revision improves stance reasoning in general, but as a coincidental alignment between the Judge's neutralising tendency and the class skew of that particular dataset.

\begin{table}[h!]                                                                                                                                                             
      \centering                                                                                                                                                                
      \resizebox{.49\textwidth}{!}{%                                                                                                                                            
      \begin{tabular}{l l r r}                                                                                                                                                  
      \toprule                                                                                                                                                                  
      Dataset & Class & share (\%) & $\Delta$F1 (pp) \\                                                                                                                         
      \midrule                                                                                                                                                                  
      \texttt{covid19-de} & \textit{positiv} & 10.3 & -4.35$_{\pm6.15}$ \\                                                                                                      
       & \textit{negativ} & 9.2 & -1.65$_{\pm15.24}$ \\                                                                                                                         
       & \textit{neutral} & 80.5 & +15.85$_{\pm11.46}$ \\                                                                                                                       
      \midrule                                                                                                                                                                  
      \texttt{covid19-glandt} & \textit{against} & 22.9 & -14.97$_{\pm18.74}$ \\                                                                                                
       & \textit{favor} & 42.9 & +2.30$_{\pm10.52}$ \\                                                                                                                          
       & \textit{none} & 34.2 & -9.06$_{\pm11.26}$ \\                                                                                                                           
      \midrule                                                                                                                                                                  
      \texttt{semeval2016t6} & \textit{against} & 44.9 & -21.25$_{\pm29.18}$ \\                                                                                                 
       & \textit{favor} & 28.3 & -10.99$_{\pm13.86}$ \\                                                                                                                         
       & \textit{none} & 26.8 & -10.96$_{\pm22.27}$ \\                                                                                                                          
      \midrule                                                                                                                                                                  
      \texttt{wtwt} & \textit{comment} & 39.2 & -8.40$_{\pm10.05}$ \\                                                                                                           
       & \textit{refute} & 9.4 & +5.70$_{\pm9.28}$ \\                                                                                                                           
       & \textit{support} & 12.8 & +0.90$_{\pm8.29}$ \\                                                                                                                         
       & \textit{unrelated} & 38.5 & +2.20$_{\pm5.97}$ \\                                                                                                                       
      \bottomrule                                                                                                                                                               
      \end{tabular}}                                                                                                                                                            
      \caption{LLM-as-a-Judge per-class F1 change relative to the zero-shot baseline (Judge $-$ ZS), in percentage points. Averaged across the four main models and three       
  seeds. \emph{Share} is the fraction of the test set with the given gold class. Positive $\Delta$F1 concentrated on the majority/neutral class indicates that the Judge acts   
  as a stance-to-neutral corrector rather than a general-purpose calibrator.}                                                                                                   
      \label{ch:finalch:tab:judge_per_class_delta}                                                                                                                              
  \end{table}

\subsubsection{Length-stratified $\Delta$F1 for \textit{Other Tweets}}
\label{app:sec:other_tweets_length} 
We test whether the length-distraction hypothesis explains the regression of \textit{Other Tweets} on \texttt{semeval2016t6} and \texttt{wtwt}. 
For each test instance we compute the total whitespace-token count of the concatenated \textit{Other Tweets} context and assign instances to per-(dataset,~setting) length quartiles Q1--Q4.
We then report the difference in macro-F1 between the \textit{Other Tweets} setting and the corresponding no-context baseline, averaged over the four main models and three seeds, per length quartile (\Cref{tab:app_ot_length}).                           

\begin{table}[t!]                                                                                                                                                             
      \small      
      \centering                                                                                                                                                                
      \begin{tabular}{llcccc}
      \toprule
      Dataset & Setting & Q1 & Q2 & Q3 & Q4 \\
      \midrule                                                                                                                                                                  
      \multirow{2}{*}{\texttt{covid19-glandt}}
          & Zero-Shot & $+4.81$ & $+1.65$ & $+2.65$ & $+6.06$ \\                                                                                                                
          & Few-Shot  & $+1.62$ & $-0.92$ & $-0.82$ & $+2.07$ \\                                                                                                                
      \midrule                                                                                                                                                                  
      \multirow{2}{*}{\texttt{semeval2016t6}}                                                                                                                                   
          & Zero-Shot & $+0.54$ & $+0.37$ & $-1.55$ & $-0.83$ \\                                                                                                                
          & Few-Shot  & $-5.08$ & $-3.46$ & $-5.44$ & $-6.53$ \\                                                                                                                
      \midrule                                                                                                                                                                  
      \multirow{2}{*}{\texttt{wtwt}}                                                                                                                                            
          & Zero-Shot & $-1.77$ & $-1.60$ & $-2.60$ & $-4.34$ \\                                                                                                                
          & Few-Shot  & $-6.80$ & $-7.05$ & $-6.69$ & $-6.58$ \\                                                                                                                
      \bottomrule                                                                                                                                                               
      \end{tabular}                                                                                                                                                             
      \caption{$\Delta$F1 (in percentage points) of \textit{Other Tweets} vs.\ the corresponding no-context baseline, stratified by per-instance length quartile of the         
  concatenated \textit{Other Tweets} context. Quartiles are computed within each (dataset, setting). Values are averaged across Ministral3-14B, Gemma3-27B, Qwen2.5-32B,         
  GPT-OSS-20B and three seeds.}
      \label{tab:app_ot_length}                                                                                                                                                 
  \end{table}
If length-induced distraction were the dominant mechanism, $\Delta$F1 should decline monotonically from Q1 to Q4.
We do not observe this pattern.
On \texttt{covid19-glandt}, the largest gains occur in both the shortest and the longest quartile in zero-shot ($+4.81$pp and $+6.06$pp), and the few-shot profile is U-shaped rather than monotonic.
On \texttt{wtwt} few-shot, $\Delta$F1 is essentially constant across length quartiles ($-6.58$ to $-7.05$pp), indicating that the regression is independent of the amount of
additional text.
Only \texttt{semeval2016t6} few-shot and \texttt{wtwt} zero-shot exhibit a mild monotonic trend, but in both cases all quartiles including the shortest show negative $\Delta$F1, suggesting that length amplifies but does not introduce the harmful effect. 
%We additionally attempted a complementary stratification by the actual number of \textit{Other Tweets} present per instance ($k_\text{actual}$); however, the strong skew toward $k_\text{actual} = 10$ ($>95\%$ of users on every dataset) renders this analysis underpowered, and we therefore do not report it here.

\subsubsection{Prediction change analysis for \textit{Other Tweets}}
\label{app:sec:othertweets_flips}

We compare \textit{Other Tweets} and no-context baseline predictions at the instance level for each (dataset, setting, model, seed) configuration to identify how often \textit{Other Tweets} introduces wrong predictions ($b_\checkmark \!\to\! o_\times$, ``regressions'') versus enables correct ones ($b_\times \!\to\! o_\checkmark$, ``gains''), and where the resulting wrong predictions concentrate.

\paragraph{Prediction change rates.}

\Cref{tab:app_ot_flip_rates} reports per-model regression and gain rates aggregated across three seeds, with the net effect (gains $-$ regressions) shown for completeness.
The regression and gain rates are large in absolute terms even when the net effect is small or zero (e.g.\ \texttt{covid19-glandt} few-shot, GPT-OSS-20B: $18.98\%$ regressions vs.\ $13.34\%$ gains for a net of $-5.64$pp), indicating that \textit{Other Tweets} substantially perturbs the prediction distribution rather than affecting only a small subset of instances.                                                              
\begin{table*}[t!]
  \small
  \centering 
  \resizebox{.99\textwidth}{!}{%
  \begin{tabular}{llcccc|cccc|cccc}
  \toprule
  & & \multicolumn{4}{c|}{Regressions ($b_\checkmark \to o_\times$)} & \multicolumn{4}{c|}{Gains ($b_\times \to o_\checkmark$)} & \multicolumn{4}{c}{Net} \\
  \cmidrule(lr){3-6} \cmidrule(lr){7-10} \cmidrule(lr){11-14}
  Dataset & Setting & Min. & Gem. & Qwen & GPT & Min. & Gem. & Qwen & GPT & Min. & Gem. & Qwen & GPT \\
  \midrule
  \multirow{2}{*}{\texttt{covid19-glandt}}
  & Zero-Shot & $9.54$  & $7.90$  & $9.64$  & $11.24$ & $18.62$ & $12.65$ & $13.50$ & $14.09$ & $+9.08$ & $+4.75$ & $+3.87$ & $+2.85$ \\
  & Few-Shot  & $8.33$  & $10.91$ & $9.70$  & $18.98$ & $17.11$ & $13.37$ & $10.95$ & $13.34$ & $+8.78$ & $+2.46$ & $+1.25$ & $-5.64$ \\                     \midrule
  \multirow{2}{*}{\texttt{semeval2016t6}}
  & Zero-Shot & $4.58$  & $14.55$ & $10.36$ & $12.15$ & $13.97$ & $5.63$  & $7.37$  & $15.83$ & $+9.39$ & $-8.93$ & $-2.99$ & $+3.69$ \\
  & Few-Shot  & $13.12$ & $16.96$ & $13.04$ & $17.15$ & $9.00$  & $9.20$  & $7.33$  & $16.22$ & $-4.11$ & $-7.76$ & $-5.70$ & $-0.93$ \\
  \midrule
  \multirow{2}{*}{\texttt{wtwt}}                                                   & Zero-Shot & $12.26$ & $9.04$  & $11.31$ & $10.94$ & $4.53$  & $7.45$  & $3.19$  & $7.74$  & $-7.73$ & $-1.59$ & $-8.12$ & $-3.20$ \\
  & Few-Shot  & $11.06$ & $15.00$ & $9.03$  & $15.89$ & $6.60$  & $6.31$  & $3.66$  & $8.10$  & $-4.46$ & $-8.69$ & $-5.37$ & $-7.79$ \\
  \bottomrule
  \end{tabular}
  }
  \caption{Per-instance prediction-flip rates (in \%) when adding \textit{Other Tweets} to the no-context baseline, aggregated across three seeds. Min.\ = Ministral3-14B, Gem.\ = Gemma3-27B, Qwen = Qwen2.5-32B, GPT = GPT-OSS-20B. Regressions and gains are reported as fractions of the entire test set; \emph{Net}\,$=$\,Gains\,$-$\,Regressions.} 
  \label{tab:app_ot_flip_rates}
\end{table*}

\paragraph{Direction of class prediction changes for \textit{Other Tweets}.}

We further inspect the class label predicted by \textit{Other Tweets} in the  cases where the output was changed to an incorrect prediction (\Cref{tab:app_ot_flip_class}).
We hypothesize that under length-induced distraction the wrong predictions would be approximately uniform across the dataset's stance classes. 
Instead, we find a strong concentration on stance-bearing labels.
On \texttt{covid19-glandt}, more than $90\%$ of regressions land on \emph{against} or \emph{favor} ($95.6\%$ in zero-shot, $90.9\%$ in few-shot), with the neutral \emph{none} class accounting for only $4.4\%$ and $9.2\%$ respectively despite being a frequent ground-truth label.                                               

On \texttt{semeval2016t6}, \emph{against} and \emph{favor} jointly account for $72$--$76\%$ of regressions.
On \texttt{wtwt}, the three stance-bearing labels (\emph{comment}, \emph{refute}, \emph{support}) account for $94.0$--$97.9\%$, leaving only $2.1$--$6.0\%$ for \emph{unrelated}.

This directional shift is consistent with a content-induced bias whereby exposure to a user's recent timeline pushes the model toward predicting that the user holds an opinionated stance on the test target, even when the correct label is neutral or unrelated.
                  
\begin{table}[t!]                                                                  \small
\centering
\begin{tabular}{llccccccc}
\toprule
Dataset & Setting & against & favor & none & comment & refute & support & unrelated \\
\midrule
\multirow{2}{*}{\texttt{covid19-glandt}} & Zero-Shot & $54.3$ & $41.3$ & $4.4$  & --     & --     & --     & --    \\
& Few-Shot  & $55.5$ & $35.4$ & $9.2$  & --     & --     & --     & --    \\ 
\midrule
\multirow{2}{*}{\texttt{semeval2016t6}} & Zero-Shot & $30.4$ & $46.0$ & $23.6$ & --     & --     & --     & --    \\
& Few-Shot  & $34.1$ & $37.9$ & $28.0$ & --     & --     & --     & --    \\
\midrule 
\multirow{2}{*}{\texttt{wtwt}} & Zero-Shot & --     & --     & --     & $36.3$ & $33.0$ & $24.7$ & $6.0$ \\ 
& Few-Shot  & --     & --     & --     & $42.8$ & $33.1$ & $22.0$ & $2.1$ \\  \bottomrule
\end{tabular}
\caption{Class distribution (in \%) of \textit{Other Tweets} predictions in the regression cell ($b_\checkmark \!\to\! o_\times$): given that the no-context baseline was correct, where does \textit{Other Tweets} now land? Aggregated across the four main models and three seeds. The label spaces of \texttt{covid19-glandt} and \texttt{semeval2016t6} (\emph{against, favor, none}) differ from \texttt{wtwt} (\emph{comment, refute, support, unrelated}); empty cells (--) denote labels not in the dataset's stance space.}
\label{tab:app_ot_flip_class}
\end{table}

\subsubsection{Prediction change analysis for \textit{Tweet-specific Context}}
\label{app:sec:tweetspecificcontext_flips}

\paragraph{Direction of class prediction changes for \textit{Tweet-specific Context}.}

In \Cref{ch:finalch:tab:tsc_regression_class_direction}, we provide a similar analysis for context \textit{Tweet-specific context} as done for \textit{Other Tweets} in \Cref{app:sec:othertweets_flips}.
Similarly, the changes in predictions are directed mostly towards non-neutral stance classes, except for the few-shot setting for \texttt{covid19-glandt} and \texttt{semeval2016t6}.

\begin{table*}[h!]
    \centering
    \resizebox{.99\textwidth}{!}{%
    \begin{tabular}{l l r r r r r r r r r r }
    \toprule
    Dataset & Setting & \textit{against} & \textit{favor} & \textit{none} & \textit{negativ} & \textit{positiv} & \textit{neutral} & \textit{comment} & \textit{refute} & \textit{support} & \textit{unrelated} \\
    \midrule
    \texttt{covid19-de} & Zero-Shot & -- & -- & -- & 52.0 & 39.7 & 8.3 & -- & -- & -- & -- \\
     & Few-Shot & -- & -- & -- & 67.6 & 28.9 & 3.5 & -- & -- & -- & -- \\
    \midrule
    \texttt{covid19-glandt} & Zero-Shot & 53.7 & 41.1 & 5.3 & -- & -- & -- & -- & -- & -- & -- \\
     & Few-Shot & 58.7 & 20.9 & 20.3 & -- & -- & -- & -- & -- & -- & -- \\
    \midrule
    \texttt{semeval2016t6} & Zero-Shot & 22.4 & 65.7 & 11.9 & -- & -- & -- & -- & -- & -- & -- \\
     & Few-Shot & 21.1 & 52.9 & 26.0 & -- & -- & -- & -- & -- & -- & -- \\
    \midrule
    \texttt{wtwt} & Zero-Shot & -- & -- & -- & -- & -- & -- & 57.1 & 19.7 & 19.2 & 4.0 \\
     & Few-Shot & -- & -- & -- & -- & -- & -- & 55.6 & 30.2 & 12.0 & 2.2 \\
    \bottomrule
    \end{tabular}}
    \caption{Class distribution (in \%) of \emph{Tweet-Specific Context} predictions in the regression cell ($b_\checkmark \to o_\times$): given that the no-context baseline was correct, where does adding \emph{Tweet-Specific Context} now land? Aggregated across the four main models and three seeds. Empty cells (\emph{--}) denote labels not in the dataset's stance space.}
    \label{ch:finalch:tab:tsc_regression_class_direction}
\end{table*}

\paragraph{LLM-generated context \textit{Tweet-Specific content} containing stance-bearing words.}

Both \textit{Target Description} and \textit{Tweet-Specific content} are LLM-generated, however, significant performance improvements are only observed for \textit{Target Description}. 
A plausible explanation for the performance regression caused by \textit{Tweet-Specific content} is that, despite the explicit instruction not to interpret the tweet, the generator provides semantic interpretations of the tweet and target, such as stance, sentiment, or rhetorical posture to the author or to the tweet itself.
This can effectively influence the downstream classifier in the stance prediction.
We test this by constructing a bilingual lexicon (English and German) of stance-attributive language, containing third-person attribution verbs (e.g.\ criticises, endorses, kritisiert, befürwortet), attitudinal predicates (is critical of, ist skeptisch), explicit attributions of mental or affective states (expresses doubt, äußert Zweifel), and markers of irony and sarcasm, and implicational frames (implying that, deutet darauf hin).

The lexicon is implemented as a set of word-bounded regular expressions and applied to the generated contexts. 
Every test instance is assigned to an attributive subset (\textit{stance-bearing}) if its context contains at least one match and to the neutral subset otherwise.
We present the rates how often a stance-bearing term was found in the following.
\begin{itemize}
    \item \texttt{covid19-de}: 1272/3046 = 41.8\%
    \item \texttt{covid19-glandt}: 408/1017 = 40.1\%
    \item \texttt{semeval2016t6}: 423/859 = 49.2\%
    \item \texttt{wtwt}: 804/8952 = 9.0\%
\end{itemize}
The match rate is substantial on the three person-centric datasets and markedly lower on \texttt{wtwt}, where targets are corporate mergers rather than individuals or claims and stance attribution to an "author" is largely inapplicable.

We then compute the rate at which the context changes a previously correct baseline prediction to an incorrect one ($\downarrow$), the symmetric rate of incorrect-to-correct changes ($\uparrow$), and the change in macro-F1 ($\Delta$F1).
Results are averaged across the three random seeds.
As \Cref{ch:finalch:tab:tsc_leakage_probe} shows, on the three person-centric datasets (\texttt{covid19-de}, \texttt{covid19-glandt}, \texttt{semeval2016t6}), the stance-bearing subgroup consistently underperforms the neutral one.
The gap persists under few-shot inference but narrows or inverts on \texttt{covid19-de} and \texttt{semeval2016t6}, where in-context demonstrations appear to dampen the effect. 
On \texttt{wtwt}, the two subgroups are indistinguishable, consistent with the low match rate and the dataset's non-attributive target structure. 
Since the two subgroups share the same model, prompt template, and seeds, and differ only in a lexical property of the generated context, the differential degradation on the person-centric datasets provides instance-level evidence that prediction performance loss due to integrating \textit{Tweet-Specific Context} is driven, at least in part, by the context-generating LLM generating stance-bearing context even though the instruction explicitly stated not to do so.

\begin{table}[h!]
    \centering
    \resizebox{.49\textwidth}{!}{%
    \begin{tabular}{l l l r r r r}
    \toprule
    Dataset & Setting & Subset & $n$ & $\downarrow$ (\%)  & $\uparrow$ (\%) & $\Delta$F1 \\
    \midrule
    \texttt{covid19-de} & Zero-Shot & stance-bearing & 1272 & 14.3 & 5.9 & -5.02 \\
     &  & clean & 1774 & 12.8 & 6.6 & -3.50 \\
     & Few-Shot & stance-bearing & 1272 & 12.3 & 7.5 & -2.41 \\
     &  & clean & 1774 & 10.9 & 7.1 & -2.52 \\
    \midrule
    \texttt{covid19-glandt} & Zero-Shot & stance-bearing & 408 & 11.8 & 14.4 & +0.69 \\
     &  & clean & 609 & 10.4 & 15.1 & +4.76 \\
     & Few-Shot & stance-bearing & 408 & 10.4 & 15.1 & +4.29 \\
     &  & clean & 609 & 7.4 & 16.8 & +9.66 \\
    \midrule
    \texttt{semeval2016t6} & Zero-Shot & stance-bearing & 423 & 14.2 & 15.3 & -0.80 \\
     &  & clean & 436 & 13.8 & 19.0 & +3.46 \\
     & Few-Shot & stance-bearing & 423 & 13.5 & 17.2 & +3.53 \\
     &  & clean & 436 & 13.8 & 15.4 & +1.40 \\
    \midrule
    \texttt{wtwt} & Zero-Shot & stance-bearing & 829 & 15.0 & 7.5 & -5.92 \\
     &  & clean & 8434 & 14.2 & 7.3 & -6.46 \\
     & Few-Shot & stance-bearing & 829 & 11.6 & 9.0 & -1.10 \\
     &  & clean & 8434 & 11.7 & 9.0 & -2.68 \\
    \bottomrule
    \end{tabular}}
    \caption{\emph{Tweet-Specific Context} prediction change rates and macro-F1 change, stratified by whether the generated context contains stance-bearing words about the author or the tweet. Computed on \texttt{GPT-OSS-20B} across three seeds. Larger $\Delta$F1 degradation on the \emph{stance-bearing} subset supports the hypothesis that the model is misled by stance and sentiment interpretation that the LLM-based context generator generates despite an explicit ``do not interpret'' instruction.}
    \label{ch:finalch:tab:tsc_leakage_probe}
\end{table}

\subsubsection{Few-shot instance-level impact of context integration}

We provide the corresponding instance-level analysis for the few-shot setting in \Cref{app:ch:finalch:tab:instance_level_comparison_gptoss20b_fewshot}.
The per-feature ratios are largely consistent with the zero-shot results. \textit{Target Description} retains a favourable improved-to-degraded ratio across all datasets, \textit{Tweet-Specific Context} preserves its high-churn signature, and \textit{Username} and \textit{User Description} remain roughly balanced.
A notable difference exists for \textit{Other Tweets} where degradation rates rise from 11--12\% (zero-shot) to 16--19\% (few-shot) on \texttt{covid19-glandt}, \texttt{semeval2016t6}, and \texttt{wtwt}, while improvement rates remain comparable (13--16\%), decisively flipping the improved-to-degraded ratio against this feature.

\begin{table}[h!]
      \centering
      \resizebox{.49\textwidth}{!}{%
      \begin{tabular}{lllll}                                                                                                                                                    
      \toprule
      Feature & covid19-de & covid19-glandt & semeval2016t6 & wtwt \\                                                                                                           
      \midrule                                                                                                                                                                  
      \multirow{3}{3cm}{\textbf{Target Description}}& {\color{green}$\uparrow$} 8\% (243) & {\color{green}$\uparrow$} 14\% (139) & {\color{green}$\uparrow$} 13\% (114) &
  {\color{green}$\uparrow$} 8\% (751) \\                                                                                                                                        
      & {\color{red}$\downarrow$} 5\% (164) & {\color{red}$\downarrow$} 7\% (67) & {\color{red}$\downarrow$} 8\% (72) & {\color{red}$\downarrow$} 9\% (828) \\                
      & {\color{gray}--} 87\% (2638) & {\color{gray}--} 80\% (810) & {\color{gray}--} 78\% (673) & {\color{gray}--} 83\% (7683) \\                                              
      \midrule                                                                                                                                                                  
      \multirow{3}{3cm}{\textbf{Username}}& {\color{green}$\uparrow$} 6\% (197) & {\color{green}$\uparrow$} 11\% (112) & {\color{green}$\uparrow$} 11\% (98) &                  
  {\color{green}$\uparrow$} 8\% (715) \\                                                                                                                                        
      & {\color{red}$\downarrow$} 7\% (200) & {\color{red}$\downarrow$} 10\% (103) & {\color{red}$\downarrow$} 11\% (91) & {\color{red}$\downarrow$} 8\% (752) \\               
      & {\color{gray}--} 87\% (2649) & {\color{gray}--} 79\% (802) & {\color{gray}--} 78\% (670) & {\color{gray}--} 84\% (7796) \\                                              
      \midrule                                                                                                                                                                  
      \multirow{3}{3cm}{\textbf{Tweet-Specific Context}}& {\color{green}$\uparrow$} 7\% (222) & {\color{green}$\uparrow$} 16\% (164) & {\color{green}$\uparrow$} 16\% (140) &   
  {\color{green}$\uparrow$} 9\% (830) \\                                                                                                                                        
      & {\color{red}$\downarrow$} 11\% (349) & {\color{red}$\downarrow$} 9\% (88) & {\color{red}$\downarrow$} 14\% (117) & {\color{red}$\downarrow$} 12\% (1081) \\           
      & {\color{gray}--} 81\% (2475) & {\color{gray}--} 75\% (765) & {\color{gray}--} 70\% (602) & {\color{gray}--} 79\% (7352) \\                                              
      \midrule                                                                                                                                                                  
      \multirow{3}{3cm}{\textbf{User Description}}& {\color{green}$\uparrow$} 8\% (249) & {\color{green}$\uparrow$} 13\% (133) & {\color{green}$\uparrow$} 14\% (118) &         
  {\color{green}$\uparrow$} 8\% (716) \\                                                                                                                                        
      & {\color{red}$\downarrow$} 6\% (192) & {\color{red}$\downarrow$} 9\% (91) & {\color{red}$\downarrow$} 11\% (93) & {\color{red}$\downarrow$} 10\% (894) \\              
      & {\color{gray}--} 86\% (2605) & {\color{gray}--} 78\% (793) & {\color{gray}--} 75\% (648) & {\color{gray}--} 83\% (7653) \\                                              
      \midrule                                                                                                                                                                  
      \multirow{3}{3cm}{\textbf{Other Tweets}}& -- & {\color{green}$\uparrow$} 13\% (136) & {\color{green}$\uparrow$} 16\% (139) & {\color{green}$\uparrow$} 8\% (750) \\       
      & -- & {\color{red}$\downarrow$} 19\% (193) & {\color{red}$\downarrow$} 17\% (147) & {\color{red}$\downarrow$} 16\% (1472) \\                                             
      & -- & {\color{gray}--} 68\% (688) & {\color{gray}--} 67\% (572) & {\color{gray}--} 76\% (7041) \\                                                                        
      \bottomrule                                                                                                                                                               
      \end{tabular}                                                                                                                                                             
      }                                                                                                                                                                         
      \caption{Instance-level impact of context integration for \texttt{GPT-OSS-20B} (few-shot), compared with the no-context baseline. Within each cell, the first row shows 
  instances where the prediction is improved (baseline wrong, context correct), the second row instances where it is degraded (baseline correct, context wrong), and the third  
  row instances where the correctness is unchanged. Counts are averaged across three seeds and rounded to the nearest integer; cells marked `--' have no runs
  (\texttt{covid19-de} has no \emph{Other Tweets} configuration).}                                                                                                              
      \label{app:ch:finalch:tab:instance_level_comparison_gptoss20b_fewshot}                                                                                                      
  \end{table}

\end{document}